\documentclass[lettersize,journal]{IEEEtran}
\usepackage{amsmath,amsfonts}

\ifCLASSOPTIONcompsoc
  \usepackage[nocompress]{cite}
\else
  \usepackage{cite}
\fi
\usepackage{orcidlink} 
\usepackage{graphicx}%
\usepackage{multirow}%
\usepackage{amsmath,amssymb,amsfonts}%
\usepackage{amsthm}%
\usepackage{mathrsfs}%
\usepackage{xcolor}%
\usepackage{textcomp}%
\usepackage{manyfoot}%
\usepackage{booktabs}%
\usepackage{algorithm}%
\usepackage{algorithmicx}%
\usepackage{algpseudocode}%
\usepackage{listings}%
\usepackage{tabularx,array,enumitem} %
\newcolumntype{L}[1]{>{\raggedright\arraybackslash}p{#1}}
\newcolumntype{Y}{>{\raggedright\arraybackslash}X}
\usepackage{float}
\usepackage{csvsimple}
\usepackage{adjustbox}
\usepackage{subcaption}
\usepackage[utf8]{inputenc}
\usepackage{amssymb}
\usepackage{threeparttable}
\DeclareUnicodeCharacter{2264}{\ensuremath{\le}}
\DeclareUnicodeCharacter{2265}{\ensuremath{\ge}}
\usepackage{placeins}
\usepackage{tabularx}
\usepackage{url}

\begin{document}

\title{Tri-Modal Severity Fused Diagnosis across Depression and Post-traumatic Stress Disorders}

\author{%
  Filippo~Cenacchi\,\orcidlink{0000-0003-2732-186X},\;
  Deborah~Richards\,\orcidlink{0000-0002-7363-1511},\;
  and~Longbing~Cao\,\orcidlink{0000-0003-1562-9429}%
  \IEEEcompsocitemizethanks{%
    \IEEEcompsocthanksitem F. Cenacchi, D. Richards, and L. Cao are with the Frontier AI Research Centre, School of Computing, Macquarie University, Sydney, NSW 2113, Australia.
    \protect\\ E-mail: \{filippo.cenacchi,deborah.richards,longbing.cao\}@mq.edu.au
  }%
  \thanks{Manuscript submitted to IEEE Transactions on Affective Computing.}%
}

\IEEEtitleabstractindextext{%
\begin{abstract}
\textbf Depression and post-traumatic stress disorder (PTSD) often co-occur with connected symptoms, complicating automated assessment, which is often binary and disorder-specific. Clinically useful diagnosis needs \emph{severity-aware} cross-disorder estimates and decision-support explanations. Our unified tri-modal affective severity framework synchronizes and fuses  interview \emph{text} with sentence-level transformer embeddings, \emph{audio} with log-Mel statistics with deltas, and \emph{facial} signals with action units, gaze, head-pose descriptors to output graded severities for diagnosing both depression (PHQ-8; 5 classes) and PTSD (3 classes).  Standardized features are fused via a calibrated late-fusion classifier, yielding per-disorder probabilities and feature-level attributions. This severity-aware tri-modal affective fusion approach is demoed on \emph{multi-disorder} concurrent depression and PTSD assessment. Stratified cross-validation on DAIC-derived corpora outperforms unimodal/ablation baselines. The fused model matches the strongest unimodal baseline on accuracy and weighted F1, while improving decision-curve utility and robustness under noisy or missing modalities. For PTSD specifically, fusion reduces regression error and improves class concordance. Errors cluster between adjacent severities; extreme classes are identified reliably. Ablations show text contributes most to depression severity,  audio–facial cues are critical for PTSD, whereas attributions align with linguistic and behavioral markers. 
Our approach offers reproducible evaluation and clinician-in-the-loop support for affective clinical decision-making.
\end{abstract}

}

\maketitle
\IEEEdisplaynontitleabstractindextext
\IEEEpeerreviewmaketitle

\markboth{IEEE Transactions on Affective Computing}{Cenacchi \MakeLowercase{\textit{et al.}}: Multi-Disorder Mental Health AI Diagnosis}

\section{Introduction}
\label{sec:introduction}
\IEEEPARstart{T}{he} World Health Organization (WHO) projects  depression as the leading cause of global disease burden by 2030 \cite{salleh2018burden}, also verified by the global COVID-19 research which identified mental health as mostly concerned during the pandemic \cite{COVID192024global}. This underscores the urgent necessity of effective interventions and policy responses. The economic implications are equally profound, with mental health conditions expected to generate costs of nearly \$6 trillion annually by 2030, driven by direct healthcare expenditures and indirect losses in productivity \cite{sowers2019mental}. Major depressive disorder (MDD) remains a leading global cause of disability, with the Global Burden of Disease study estimating 274.8 million new cases in 2019, representing a 59\% increase since 1990 \cite{yan2024global}. The prevalence and impact of post-traumatic stress disorder (PTSD) have intensified globally during and after the COVID-19 pandemic; meta-analytic evidence estimates a pooled prevalence of 19.34\% among the general population, with no significant regional variation \cite{hoang2023prevalence}. Healthcare workers, exposed to persistent occupational stressors, reported prevalence rates of 13.3\% in countries such as Spain, Italy, and the United States \cite{abdullah2022prevalence}. In China, prevalence reached 25.2\% one month after the outbreak, with elevated rates among individuals directly affected by pandemic-related stressors \cite{lei2021prevalence}. Beyond heightened prevalence, PTSD has been consistently linked to cognitive impairment, sleep disturbances, comorbid depression, and elevated mortality risk, with major risk factors including female gender, prior anxiety or depression, and bereavement-related stress \cite{field2024post}.

PTSD and depression frequently co-occur, with depressive symptoms exerting a profound negative impact on quality of life (QoL) in individuals diagnosed with PTSD. Among shared symptoms, anhedonia emerges as a particularly salient predictor of diminished QoL, underscoring the interwoven psychopathology of these conditions \cite{miller2024quality}. Standard diagnostic methods, including structured interviews and self-report questionnaires, remain indispensable in clinical practice but are constrained by subjectivity and vulnerability to reporting biases. These limitations have motivated the exploration of complementary computational approaches capable of capturing objective behavioral signals in naturalistic settings. The Patient Health Questionnaire-8 (PHQ-8) is one of the most widely used instruments for assessing depression severity and is frequently employed alongside PTSD-specific measures such as the PTSD Checklist for DSM-5 (PCL-5) to evaluate symptom severity and treatment outcomes \cite{kline2024ptsd, brigido2021posttraumatic}. 

Epidemiological and clinical evidence consistently shows that PTSD and depression frequently co-occur, with comorbidity associated with greater symptom severity, reduced quality of life, and elevated suicide risk. In war-exposed populations, such co-occurrence significantly amplifies distress and functional impairment \cite{pejuskovic2020longitudinal}. Among active-duty service members, latent profile analyses have identified distinct subgroups based on symptom patterns of PTSD and depression, each displaying differential treatment responsiveness and highlighting the complexity of their interaction \cite{kline2024ptsd}. While instruments such as the PHQ-8 and PCL-5 provide validated severity estimates, they remain resource-intensive and may not capture the full heterogeneity of symptom presentation, necessitating the development of complementary approaches that transcend reliance on subjective self-report \cite{radell2020depression}. In this context, computational psychiatry has increasingly turned toward computer-based measures that can objectively quantify behavioral signals linked to PTSD and depression. These methods not only offer the potential to reduce diagnostic subjectivity but also bridge findings from human and animal research, providing mechanistic insight into the biological substrates of these disorders and informing the design of novel interventions \cite{radell2020depression}. The clinical implications of comorbidity are profound, as the severity of PTSD symptoms can influence the effectiveness of treatments for depression. For example, while PTSD diagnosis alone does not predict treatment outcomes, heightened PTSD symptom severity has been shown to affect the clinical response and remission rates of depression interventions such as transcranial magnetic stimulation (TMS) \cite{brigido2021posttraumatic}. Evidence-based therapies for PTSD, including cognitive–behavioral interventions and pharmacological treatments, remain recommended as first-line options, yet the presence of comorbid depression frequently necessitates tailored and multimodal strategies to achieve optimal outcomes \cite{pejuvskovic2024posttraumatic}. Together, these findings illustrate the urgent need for diagnostic paradigms that move beyond traditional self-report and interview-based methods. Computational approaches provide a promising avenue for developing objective, scalable, and clinically relevant tools to capture the intertwined pathology of PTSD and depression, thereby enhancing diagnostic precision and improving treatment personalization.

\medskip
  Recent advances in artificial intelligence (AI) have significantly expanded the potential for automated assessment of depression and PTSD by leveraging linguistic, acoustic, and visual cues. These modalities encode rich behavioral information that can be systematically analyzed to infer underlying mental states, supporting both screening and longitudinal monitoring. AI models are now capable of extracting discriminative features from speech prosody, lexical content, and facial dynamics, providing promising alternatives to traditional clinical assessments, which are often resource-intensive and subject to reporting biases. One line of research has emphasized the diagnostic value of visual and auditory markers. Schultebraucks et al.\ \cite{schultebraucks2022deep} demonstrated that deep learning models could accurately classify PTSD and MDD from free speech responses, reporting an AUC of 0.90 for PTSD and 0.86 for depression . Building on this, Uddin et al.\ \cite{uddin2022deep} proposed a deep multimodal framework that integrates facial and verbal cues for depression assessment, which outperformed unimodal approaches . Parallel work has shown that even unimodal channels can yield strong predictive signals: König et al.\ \cite{konig2022detecting} reported that automated speech analysis in a non-clinical sample could predict depression scores with high accuracy. Complementary findings have emerged in PTSD detection, where Quatieri et al.\ \cite{quatieri2023emotion} developed an emotion-driven screening tool based on vocal biomarkers, achieving an AUC of 0.80. These studies highlight the promise of AI-driven multimodal analysis for enhancing diagnostic accuracy and reducing dependence on subjective self-report, though challenges around dataset diversity, overlapping symptomatology, and ethical deployment remain \cite{dhelim2023artificial}.

\medskip
  These lines of research collectively show that (i) multimodality improves detection versus unimodal baselines, (ii) modern sequence encoders can leverage temporal structure in speech and facial behavior, and (iii) careful fusion benefits robustness. Yet, two gaps remain for translational impact: \emph{multi-disorder modeling} (capturing overlapping, interacting symptom processes across conditions), and \emph{severity-aware prediction} (moving beyond dichotomies to clinically interpretable grades aligned with validated instruments). The first gap matters because comorbidity is common and presentations are heterogeneous; the second matters because care planning depends on severity estimates.

\medskip
  \textbf{This paper addresses both gaps.} We present a \emph{tri-modal} AI diagnosis framework that jointly models depression and PTSD and produces \emph{graded} severity outputs. Our approach fuses complementary information from \emph{text}, \emph{audio}, and \emph{face} using computationally efficient feature extractors (sentence-level embeddings, log-Mel statistics with deltas, and OpenFace-derived descriptors) and a calibrated late-fusion classifier. Unlike prior disorder-specific models, we explicitly formulate \emph{multi-disorder, multi-severity} prediction and validate across corpora derived from clinical interviews, demonstrating that multimodal fusion improves discrimination and supports severity mapping aligned with PHQ-8 and PTSD categories.

\medskip
  \textbf{Contributions.} To the best of our knowledge, this work is the first to bring together \emph{tri-modal fusion}, \emph{multi-disorder coverage}, and \emph{graded severity prediction} in one practical framework. Specifically, we contribute:
\begin{itemize}
  \item \textbf{Unified multi-disorder modeling}:  Simultaneously predicts depression and PTSD, addressing real-world co-occurring misorders \cite{al2018detecting,lin2020towards,lam2019context,jo2022diagnosis}.
  \item \textbf{Graded severity estimation}: Catering for multiple rather than binary levels for respective misorders (e.g., PHQ-8 levels for depression; categorical PTSD severity).
  \item \textbf{Efficient tri-modal fusion}: Lightweight, reproducible fusion of text–audio–face signals, delivering strong discrimination without massive end-to-end transformers.
\end{itemize}

Our approach \emph{generalizes cross-disorders} and \emph{operates on severity scales}, bridging existing strands, where depression-focused multimodal systems show fusion gains but are constrained to binary classification \cite{zhang2024multimodal, chen2024iifdd, li2024fpt}; while PTSD-focused multimodal transformers advance architectural sophistication within that single diagnosis \cite{dia2024paying}. 
Pooled evaluation  demonstrates robustness of our approach on the Extended Distress Analysis Interview Corpus (E-DAIC) \cite{gratch2014distress,ringeval2019avec} and DAIC-WOZ (Distress Analysis Interview Corpus under a ``Wizard of Oz'' condition). Comprehensive results are evaluated by per-class F1, pooled confusion matrices, and ROC/PR curves.


\medskip
The paper is organized as follows. Section~\ref{sec:background} reviews related work. Section~\ref{sec:methods} details the framework. Section~\ref{sec:results} reports results. Section~\ref{sec:Discussion} discusses implications and limitations. Section~\ref{sec:Conclusion} concludes.

\section{Related Work}
\label{sec:background}

\subsection{Automated Diagnosis of Mental Health Disorders}
AI applications in mental health increasingly leverage heterogeneous data sources, ranging from electronic health records and mood rating scales to brain imaging and social media streams \cite{graham2019artificial}. Machine learning and natural language processing have emerged as predominant methodological approaches, supporting tasks such as automated symptom extraction, severity classification, and comparative evaluation of therapeutic effectiveness \cite{Gerantia_2024}. Despite these advances, significant challenges remain, including issues of data quality, algorithmic bias, and broader ethical concerns, all of which underscore the need for further research to improve diagnostic reliability and ensure equitable healthcare delivery \cite{Gerantia_2024, zhang2024can}. While AI demonstrates substantial promise in augmenting mental health assessment and treatment, its integration must be approached with caution. Ethical imperatives such as patient privacy, informed consent, and mitigation of algorithmic bias are central to ensuring that AI complements rather than supplants the human dimensions of psychotherapy. In this light, AI should be understood as an augmentative tool designed to extend clinical capacity in resource-limited contexts, while preserving clinician oversight and the human-to-human therapeutic alliance—namely, context-sensitive judgement and genuine empathic engagement—which AI systems may simulate but must not replace \cite{zhang2024can}.

Traditional approaches to diagnosing depression and PTSD, including structured interviews and self-report questionnaires, face persistent methodological and practical limitations that compromise their reliability and scalability. These methods frequently depend on subjective self-assessments, making them vulnerable to biases that can distort diagnostic accuracy. For instance, self-report questionnaires, although widely employed for initial screening, are prone to false positives as they are designed to capture a broad spectrum of symptoms, often resulting in inflated prevalence estimates \cite{zimmerman2024value}. Structured interviews anchored in International Classification of Diseases (ICD) or Diagnostic and Statistical Manual of Mental Disorders (DSM) criteria similarly fall short in capturing nuanced differential diagnoses, with vague criteria and the presence of somatic comorbidities further complicating accuracy \cite{linden2012standardized}. The challenges are amplified in populations with intellectual disabilities, where item content, phrasing, and response formats in questionnaires can undermine the validity of assessments \cite{finlay2001methodological}.  

Beyond subjectivity, scalability represents a critical obstacle. Because positive questionnaire screenings generally necessitate follow-up interviews, these tools remain resource-intensive and reliant on trained clinicians, thereby limiting applicability in large-scale settings \cite{zimmerman2024value}. Even widely used self-report scales such as the PTSD Checklist exhibit variable diagnostic performance depending on the cutpoints applied, further complicating their use in epidemiological studies \cite{magruder2015diagnostic}. Inconsistencies in symptom identification add another layer of complexity: studies of major depression demonstrate considerable variation in symptom structures across populations, rendering universal screening instruments elusive \cite{north2021symptom}. Similarly, investigations of postdisaster depression reveal heterogeneous symptom patterns that necessitate comprehensive diagnostic assessments rather than reliance on brief self-reports \cite{north2021symptom}.  

Despite these limitations, traditional diagnostic methods remain indispensable in clinical practice. Their value lies in being embedded within a broader diagnostic process that integrates initial screening with detailed follow-up evaluations, thereby mitigating the impact of reporting biases, improving specificity, and partially addressing scalability challenges. This integrated approach underscores the continued clinical utility of structured interviews and questionnaires while acknowledging the urgent need for methodological innovation to complement their use.

Speech and language characteristics have emerged as promising biomarkers for MDD, offering an avenue for objective and scalable assessment. Recent advances in automated speech analysis demonstrate the ability to distinguish depressed patients from healthy controls by identifying subtle alterations in prosody, rhythm, and articulation patterns, thereby enabling the detection of symptom severity with greater precision than self-reports or clinician impressions alone \cite{menne2024voice}. Building on this, computational models have begun to operationalize these speech-derived markers into diagnostic pipelines. For example, a deep tensor-based framework for automatic depression recognition leverages shifts in speech utterances to achieve high classification accuracy and robust F1 scores, while being lightweight enough for deployment in wearable systems that support real-time monitoring \cite{pandey2022deep}. These developments highlight the potential of speech-based diagnostics to complement traditional methods by offering continuous, unobtrusive, and context-sensitive insights into mental health trajectories, thereby addressing limitations of subjectivity and scalability that characterize conventional approaches.

\subsection{Multimodal Approaches for Depression Diagnosis}

Research on automated depression detection has expanded substantially with the availability of datasets such as DAIC-WOZ and its extension E-DAIC, which provide synchronized audio, video, and textual interview data with PHQ-8 labels. Early work often focused on unimodal signals. For example, Zhang et al.\ \cite{zhang2022_fusion_eeg} demonstrated that spatial--temporal feature-level fusion of pervasive EEG recordings improves recognition rates over single-channel analysis, while Yousufi et al.\ \cite{yousufi2024_eeg_audio} combined EEG with audio spectrograms, showing gains compared to unimodal baselines.

Building on these efforts, multiple studies have directly targeted multimodal clinical interview corpora. Wei et al.\ \cite{wei2022_subattn} introduced a sub-attention fusion network on DAIC-WOZ that integrated text, audio, and video streams, achieving precision close to 0.89 for major depression detection and reducing severity estimation error (MAE $\approx$ 4.9) compared to earlier baselines. Nykoniuk et al.\ \cite{nykoniuk2024} explored multimodal early vs.\ late fusion of audio and transcripts on DAIC-WOZ and E-DAIC, with their best early-fusion model reaching 0.86 accuracy and F1 = 0.79. Sadeghi et al.\ \cite{sadeghi2024} further leveraged E-DAIC interviews for fully automated severity prediction pipelines, reporting that multimodal models significantly outperformed text-only or audio-only counterparts in estimating PHQ-8 scores.

Several recent works refined fusion architectures. Xu et al.\ \cite{xu2025} fused BERT-based textual embeddings with Wav2Vec2-based voice features on DAIC-WOZ, yielding consistent F1 improvements (+0.06 over voice-only, +0.25 over text-only) and lowering RMSE by nearly 2 points for severity estimation. The HiQuE framework \cite{hiquE2024} modeled the hierarchical structure of DAIC-WOZ interviews, distinguishing primary from follow-up questions and embedding them with cross-modal attention; this question-aware approach further boosted accuracy relative to flat fusion models. 

Together, these studies confirm the importance of multimodality and advanced fusion strategies: while unimodal EEG or audio/text pipelines achieve solid results, leveraging synchronized modalities in DAIC-WOZ and E-DAIC interviews consistently yields superior detection accuracy and more reliable severity estimates, underscoring multimodality as the de facto standard for clinically relevant depression detection research.

The availability of large-scale multimodal datasets has been instrumental in advancing automated mental health assessment. Among these, the DAIC-WOZ and the E-DAIC \cite{gratch2014distress, ringeval2019avec} are particularly influential, providing structured data for hundreds of patients with both PHQ-8 depression scores and PTSD severity labels. By including synchronized modalities such as text, audio, and video, these corpora support the development of robust models capable of capturing the nuanced manifestations of depression and related conditions across behavioral, linguistic, and paralinguistic dimensions.

A key strength of these datasets lies in their ability to facilitate multimodal exploration. Researchers can leverage the interplay between verbal and non-verbal cues to construct richer predictive models, a direction that has proven essential given the heterogeneity of depressive symptoms \cite{bucur2023s}. Furthermore, DAIC-WOZ and E-DAIC serve as benchmark datasets that enable standardized training and evaluation of diverse architectures, ranging from LSTM-based pipelines to multimodal transformers, ensuring that methodological advances can be compared on a common ground \cite{prama2024ai, bucur2023s}. These benchmarks have catalyzed the design of time-aware, context-sensitive models that achieve higher accuracy and reliability in depression detection.

Beyond benchmarking, the datasets also enable systematic feature extraction, including emotional cues, topical shifts, and behavioral markers such as speech hesitations or gaze aversions. Such features contribute to identifying depression-related patterns that may be imperceptible in unimodal data streams \cite{prama2024ai}. Nevertheless, while DAIC-WOZ and E-DAIC have become cornerstones of multimodal affective computing research, there remains a scarcity of publicly available corpora for digital phenotyping of mental health. Existing resources are limited in scope, often underrepresenting diverse populations and interaction contexts. Recent reviews emphasize the urgent need for more comprehensive datasets that incorporate longitudinal and real-world behavioral signals, particularly through sensing applications and ecological momentary assessments, to improve generalizability and ecological validity \cite{mendes2022sensing}.

\subsection{PTSD Detection with Machine and Deep Learning}
The detection of post-traumatic stress disorder (PTSD) using machine learning and deep learning has gained increasing attention, although it remains less extensively explored compared to depression. Recent research has leveraged multi-dimensional data and advanced neural architectures to improve accuracy and reliability, with results that suggest strong clinical potential. Machine learning models have demonstrated considerable effectiveness in PTSD classification. A recent meta-analysis reports overall accuracies of 0.89, with multimodal approaches (acoustic, textual, visual) outperforming unimodal ones (0.96 vs.\ 0.86--0.90) \cite{wang2024application}. Ensemble models such as random forests have also been used to identify undiagnosed PTSD in large civilian populations, showing the potential of ML to uncover hidden cases and improve treatment outcomes \cite{gagnon2022identifying}.

Deep learning has further advanced PTSD detection by capturing both short- and long-range dependencies in multimodal data. Hybrid architectures such as CNN--LSTM and BiGRU autoencoders have been used to analyze speech emotion and extract long-term speech features for classification \cite{schultebraucks2022deep}. More recently, stochastic multimodal Transformer models trained on text, audio, and video have achieved strong results, with a concordance correlation coefficient of 0.722, demonstrating robust representational learning for PTSD symptoms \cite{dia2024paying}.

Multimodal approaches combining visual and auditory markers have reported high discriminatory accuracy, with AUCs up to 0.90 \cite{schultebraucks2022deep}. Beyond technical integration, interdisciplinary studies combining psychiatry, linguistics, and natural language processing have identified reliable linguistic markers of PTSD, achieving diagnostic performance comparable to conventional methods \cite{quillivic2024interdisciplinary}.

\subsection{Challenges and Considerations}
Despite these advances, challenges remain. PTSD frequently co-occurs with depression, complicating differential diagnosis and reducing interpretability of models. Optimal combinations of modalities for clinical deployment are still unclear, and ethical considerations such as transparency, acceptance, and trust are central to integration into real-world psychiatric practice.

\subsection{Limitations of Current Approaches}
The current literature on automated mental health assessment reveals three key limitations:
\begin{enumerate}
  \item \textbf{Single-disorder focus:} Most studies target either depression or PTSD independently, neglecting their frequent co-occurrence and interactive symptomatology.
  \item \textbf{Binary or limited outcomes:} Many systems reduce prediction to binary presence/absence or approximate regression, overlooking the graded severity levels required for clinical utility.
  \item \textbf{Fragile multimodal fusion:} Although multimodality is widely acknowledged to improve performance, existing fusion strategies often lack robustness under missing or noisy modalities, limiting deployment in real-world, resource-variable contexts.
\end{enumerate}

\section{Proposed approach}
\label{sec:methods}

\subsection{System Overview}
\label{sec:system-overview}

\begin{figure*}[t]
  \centering
  \includegraphics[width=\textwidth,clip,trim=2pt 2pt 2pt 2pt]{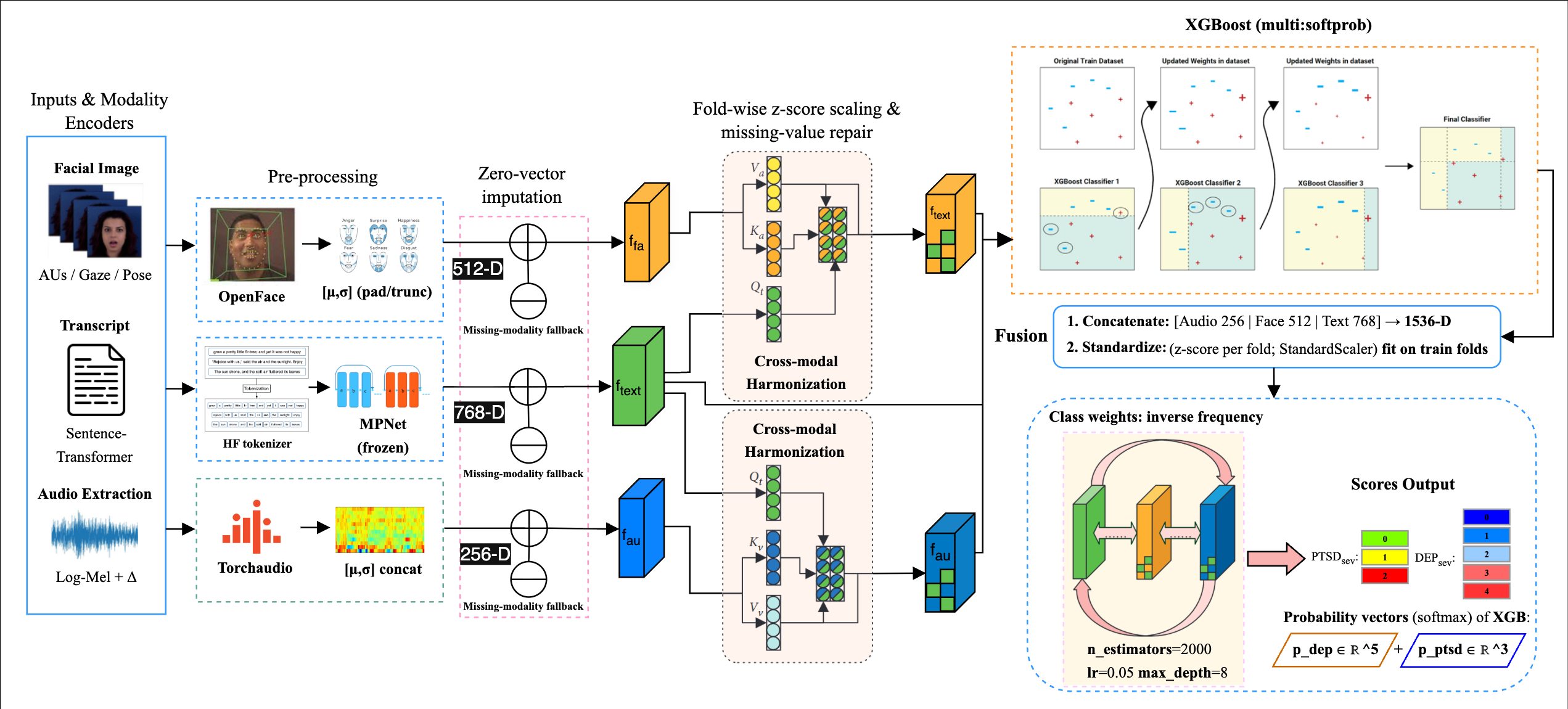}
  \caption{\textbf{Architecture overview.} Text (768-D), audio (256-D), and face (512-D) embeddings are
  standardized and fused (1,536-D) for boosted late-fusion classification (Depression, PTSD). Calibrated
  probabilities enable severity-aware diagnosis.}
  \label{fig:overview}
\end{figure*}

Our framework is a \emph{tri-modal, multi-task} pipeline that ingests synchronized \textbf{text}, \textbf{audio}, and \textbf{face} signals from clinical interviews and returns \emph{graded} severity estimates for both depression (PHQ-8 classes 0--4) and PTSD (3 classes). Figure~\ref{fig:overview} summarizes the end-to-end flow. The design emphasizes (i) \textbf{complementarity}---each modality contributes a distinct slice of information; (ii) \textbf{robustness} under partial or noisy inputs via late fusion with calibrated probabilities; and (iii) \textbf{practical reproducibility} through compact, well-specified feature sets and a strong classical learner.

\medskip
 
\textbf{Top–down dataflow (Fig.~\ref{fig:overview}).} Interviews are first cleaned to remove interviewer prompts and non-speech segments. Each stream follows a lightweight extractor tailored to clinically salient cues: (1) \emph{Text} uses sentence-level transformer embeddings to capture semantics and pragmatics, producing a 768-D vector per participant; (2) \emph{Audio} summarizes log-Mel energy distributions and their deltas (prosody, rhythm, energy shifts) into a 256-D descriptor; (3) \emph{Face} aggregates OpenFace Action Units, gaze, and head pose statistics into a 512-D vector that reflects expressivity and motoric patterns. Modality vectors are standardized on the training split, concatenated into a 1536-D fusion vector, and fed to a \emph{boosted late-fusion} classifier (\texttt{multi:softprob}) that outputs calibrated class probabilities \emph{per disorder}. We train separate heads for PHQ-8 (5-way) and PTSD (3-way), enabling disorder-specific decision boundaries while sharing the same fused representation.

\medskip
 
\textbf{Rationale.} Late fusion offers two practical advantages in this setting: (i) it tolerates \emph{heterogeneous noise/missingness}—if one stream is unreliable for a participant, the others can dominate the decision; and (ii) it produces \emph{well-behaved probabilities}, which we use downstream for threshold-sensitive clinical analyses (e.g., decision curves in Section~\ref{sec:Discussion}). The specific feature dimensions in Table~\ref{tab:exp_setup} are chosen to balance coverage and tractability; together they yield a compact 1{,}536-D space that performs strongly in cross-validated evaluation while remaining easy to reproduce.

\subsection{Text Feature Extraction}
\label{sec:text}

Textual signals represent one of the most diagnostically informative modalities in clinical interviews, as language directly conveys affect, cognition, and symptom-related discourse markers. Our design of the text pipeline therefore emphasizes both \emph{semantic depth} (capturing nuanced meaning and context) and \emph{structural compactness} (yielding reproducible, low-dimensional descriptors compatible with multimodal fusion). This choice reflects prior evidence that linguistic markers are often the strongest predictors of depression and PTSD, while also ensuring tractability when aligned with audio and visual channels.

\medskip
 
\textbf{Preprocessing.} Interview transcripts are first filtered to remove non-patient turns (e.g., interviewer prompts and system instructions), ensuring that embeddings reflect only participant speech. Standard normalization follows: lowercasing, punctuation stripping, and whitespace correction. Unlike conventional clinical text corpora, which may be heavily edited or written, DAIC-WOZ and E-DAIC transcripts preserve disfluencies, pauses, and colloquial expressions. We retain these elements, as they carry diagnostic value for both disorders (e.g., hesitations as markers of avoidance in PTSD; reduced lexical diversity in depression).

\medskip
 
\textbf{Embedding model.} Each sentence is encoded using the transformer-based model \texttt{all-mpnet-base-v2} \cite{song2020mpnet,reimers2019sentencebert}, chosen for its balance of efficiency and contextual coverage. MPNet leverages permuted language modeling with attention mechanisms that capture both syntactic dependencies and long-range discourse structure, outperforming BERT-like baselines in semantic similarity benchmarks. Sentence-level embeddings are obtained via mean pooling of token hidden states, yielding a dense 768-dimensional vector per subject. This pooling strategy has two key benefits: (i) it provides a stable, global descriptor across variable-length transcripts, and (ii) it reduces variance compared to CLS-token representations, which can be unstable on clinical conversational text.

\medskip
 
\textbf{Aggregation.} To derive participant-level representations, all sentence embeddings within an interview are averaged, producing a single 768-D vector per subject. This simple yet robust aggregation captures overall semantic tendencies across the interaction, consistent with prior work showing that mean-pooled embeddings preserve clinically relevant traits without overfitting to local idiosyncrasies. For robustness checks, we also experimented with TF-IDF bag-of-words vectors concatenated to the transformer embedding; however, these offered marginal gains while substantially increasing dimensionality, so were excluded from the final fusion space (see Table~\ref{tab:exp_setup}).

\medskip
 
\textbf{Rationale and positioning.} Compared to traditional approaches such as TF-IDF, LIWC, or static embeddings (e.g., word2vec, GloVe), our sentence-transformer strategy offers three main advantages: (i) it encodes both lexical choice and pragmatic structure, which is critical for capturing negative affect, avoidance, or trauma-narrative coherence; (ii) it yields fixed-length descriptors amenable to multimodal concatenation with audio (256-D) and face (512-D) features, aligning exactly with the 1,536-D fusion space; and (iii) it maintains computational efficiency, avoiding the prohibitive overhead of full sequence-to-sequence models (e.g., BERT fine-tuning) while still leveraging pretrained contextual knowledge. In this sense, our design aligns with and extends prior depression-only pipelines that relied on co-attention across word-level features, while also contrasting with PTSD-specific stochastic transformers that operate frame-by-frame. Instead, we prioritize reproducibility, tractability, and robustness for the multi-disorder, multi-severity setting.

\subsection{Audio Feature Extraction}
\label{sec:audio}

Vocal prosody provides a direct, non-verbal window into affective state, motor control, and cognitive load. Depressed individuals often exhibit monotonic speech, reduced energy, and slower articulation, while PTSD is characterized by heightened arousal, jitter, and abrupt changes in pitch and loudness. Capturing these dynamics requires feature representations that are both \emph{time–frequency aware} and \emph{robust to noise}. We therefore adopt log-Mel spectrogram statistics, a standard in speech emotion and paralinguistic analysis, as the backbone of the audio subsystem.

\begin{figure}[t]
  \centering
  \includegraphics[width=0.50\linewidth]{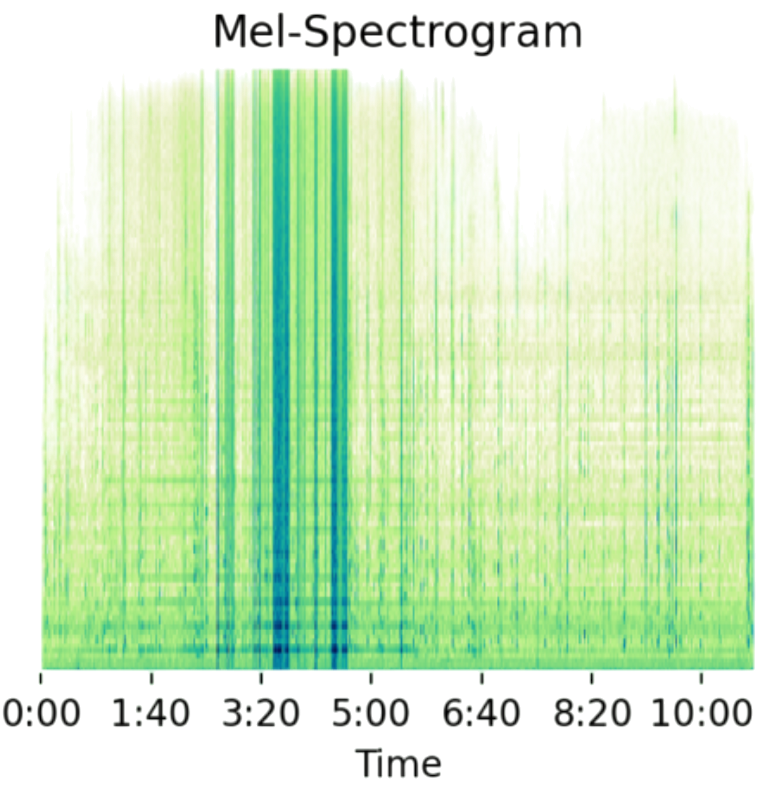}
  \caption{\textbf{Representative log-Mel spectrogram.} Example extracted at 16\,kHz with $n_{\text{mels}}=64$. Bright vertical stripes correspond to voiced segments with strong harmonic energy, while dark gaps reflect silences and low-energy speech.}
  \label{fig:audio_logmel}
\end{figure}

\medskip

\textbf{Spectro-temporal representation.} Each raw waveform is first resampled to 16\,kHz, then transformed into a Mel-scaled spectrogram with $n_{\text{mels}}=64$, a 25\,ms window, and 10\,ms stride. This yields a compact two-dimensional time–frequency map that emphasizes perceptually relevant bands while discarding redundant spectral resolution. Figure~\ref{fig:audio_logmel} shows a representative log-Mel spectrogram segment. Brighter vertical structures correspond to voiced speech bursts with strong harmonic content, while darker regions indicate silence or low-energy pauses. The temporal density of these bursts (e.g., around 3--4 minutes in the figure) provides a first-order signal of speech activity and rhythm, features often disrupted in clinical populations.

\medskip

\textbf{Statistical pooling.} Rather than feeding raw spectrogram frames into a deep sequence model (which risks overfitting in limited-data clinical settings), we summarize each recording using second-order statistics. Specifically, we compute the \emph{mean} and \emph{standard deviation} across time for each Mel bin and its first-order derivative ($\Delta$), yielding a 256-dimensional descriptor per participant. This approach preserves distributional properties of energy and prosodic dynamics while discarding local noise and session artifacts. It is also computationally efficient and reproducible, aligning with our reproducibility-first design.

\textbf{Normalization and fusion readiness.} Features are standardized within each training fold using \texttt{StandardScaler} to ensure comparability across participants. The resulting 256-D vector is concatenated with text (768-D) and face (512-D) features, contributing a balanced but complementary role in the 1,536-D fusion space. This design ensures that audio is not overwhelmed by high-dimensional linguistic representations, yet still contributes prosodic and energetic cues that text cannot capture.

\medskip

\textbf{Rationale.} The log-Mel statistical embedding strikes an optimal balance between fidelity and tractability. While end-to-end spectrogram transformers have been proposed for PTSD detection, they typically require substantially larger datasets and compute than available in DAIC-derived corpora. Our approach, in contrast, distills clinically relevant prosodic cues into a robust vector space, avoids overfitting, and integrates seamlessly with the late-fusion classifier. As later SHAP analysis shows, audio dimensions contribute substantially to model confidence in moderate-to-severe PTSD, where vocal arousal and disfluencies dominate.

\subsection{Facial Feature Extraction}

Facial dynamics constitute a primary channel for the externalization of affective states, and they offer critical diagnostic cues for depression detection. In this work, we adopted an automated feature extraction pipeline that transforms raw video streams into high-dimensional, temporally aligned descriptors of facial muscle activity. Unlike handcrafted descriptors, this approach leverages state-of-the-art computer vision models to capture both static morphological traits and dynamic micro-expressions, ensuring that subtle behavioral markers associated with depressive symptomatology are not overlooked.

\medskip

The pipeline is built on the OpenFace 2.0 framework, which provides frame-level estimates of facial landmarks, action units (AUs), head pose, and eye gaze vectors. Each video is first decomposed into individual frames at 30 fps. For every frame, 68 facial landmarks are localized and normalized to a canonical reference frame, enabling robustness against variations in camera position and head orientation. From these landmarks, shape-based geometric features are derived, while AU intensity values encode muscle activation patterns linked to emotional valence and arousal. Additionally, gaze direction and blink frequency are extracted as indicators of attentional engagement and psychomotor retardation factors frequently impaired in depression.

\medskip
 
To construct a temporally stable feature representation, we aggregated frame-level outputs across the duration of each session. For geometric features and AUs, we computed both statistical descriptors (mean, variance, skewness) and dynamic properties such as derivative-based measures of velocity and acceleration. This dual encoding preserved both the overall distribution of expressions and their rapid fluctuations, which prior studies \cite{mollahosseini2019, wang2024cmpb} have highlighted as essential markers for affective disorders. Eye-gaze and blink-based measures were similarly aggregated into frequency-domain descriptors, yielding a multimodal vector that combines spatial, temporal, and frequency features.

\medskip
 
The resulting facial embedding is a 512-dimensional vector per subject, which balances descriptive richness with computational efficiency. This representation is subsequently synchronized with audio and textual embeddings to enable late fusion across modalities. By anchoring the extraction process in validated psychometric constructs such as flattened affect, reduced expressivity, and gaze aversion, the pipeline ensures that the features are not only statistically robust but also clinically interpretable. 

\subsection{Multimodal Fusion}

The central objective of the proposed framework is to effectively integrate heterogeneous information streams—speech, facial expressions, and textual transcripts—into a coherent diagnostic representation capable of supporting robust depression and PTSD classification. While unimodal models can capture salient cues in isolation, the underlying psychopathological markers are inherently multimodal, emerging through the co-articulation of vocal prosody, micro-expressions, and language semantics. Consequently, we standardize each stream and then concatenate them into a joint vector that the classifier can exploit, yielding a robust shared representation without leakage across modalities.

\medskip
 
Our fusion strategy is probabilistic late fusion via concatenation. Each modality is embedded independently—audio (log-Mel statistics with $\Delta$; 256-D), face (OpenFace AU/gaze/pose stats; 512-D), and text (MPNet sentence embeddings; 768-D). Within each training fold, we fit a \emph{StandardScaler} on the training subset and apply it to validation data only (no PCA), then concatenate the three standardized vectors into a 1{,}536-D fused representation.

\medskip
 
We learn class probabilities with \emph{XGBoost} multi-class heads (\texttt{objective=multi:softprob}) \cite{chen2016xgboost}, one per disorder (PHQ-8: 5 classes; PTSD: 3 classes). This preserves disorder-specific decision boundaries while sharing the same fused representation. SHAP analysis \cite{lundberg2017unified} on the fitted models provides feature- and modality-level attributions, and the late-fusion design affords graceful degradation under noisy or missing streams.

\medskip
 
\textbf{Formalization.}
Let $x_a\in\mathbb{R}^{256}$, $x_f\in\mathbb{R}^{512}$, and $x_t\in\mathbb{R}^{768}$ denote the audio, face, and text embeddings. Let $\mathcal{S}$ be the standardizer fitted on the training split of each fold. The fused vector is
\begin{equation}
x \;=\; \big[\; \mathcal{S}(x_a) \; \| \; \mathcal{S}(x_f) \; \| \; \mathcal{S}(x_t) \;\big] \in \mathbb{R}^{1536}.
\end{equation}
For a task with $K$ classes (Depression: $K{=}5$; PTSD: $K{=}3$), XGBoost learns class margins
\begin{equation}
F_k(x) \;=\; \sum_{b=1}^{B} f_{b,k}(x), \qquad k=1,\ldots,K,
\end{equation}
where $f_{b,k}$ is the $b$-th regression tree for class $k$. Class probabilities are the softmax of the margins:
\begin{equation}
p_k(x) \;=\; \frac{\exp\!\big(F_k(x)\big)}{\sum_{j=1}^{K}\exp\!\big(F_j(x)\big)}.
\end{equation}
We minimize class-weighted cross-entropy
\begin{equation}
\mathcal{L} \;=\; -\frac{1}{N}\sum_{i=1}^{N} w_{y_i}\,\log p_{y_i}(x_i),
\end{equation}
with inverse-frequency weights $w_{y}$ computed per training fold. This yields calibrated, disorder-specific probabilities suitable for decision-curve analysis and per-feature attributions via SHAP.

\subsection{Diagnosis Severity Categorization}
 
The diagnosis output of the system is the following: a Depression severity score that is discretized according to validated PHQ-8 thresholds (none, mild, moderate, moderately severe, severe) \cite{kroenke2009phq8}, and a PTSD severity categorized into three levels (none/mild, moderate, severe) using the PCL-5 (DSM-5) total score (20 items scored 0–4; range 0–80): \(\leq 20\) none/mild, \(21\text{–}40\) moderate, and \(\geq 41\) severe \cite{blevins2015pcl5}. This three-band mapping is a simple binning for interpretability aligned to our dataset distribution; continuous analyses use the 0–80 totals. By predicting categorical severity, our framework aligns with clinical practice, supports longitudinal monitoring, and provides finer-grained insights into patient status.

\subsection{Training and Evaluation Protocol}
 
The system is evaluated on both the DAIC-WOZ and E-DAIC corpora to demonstrate robustness across distinct clinical interview settings. To mitigate data leakage, splits are performed at the participant level rather than the session or utterance level. Each task—depression and PTSD classification—is assessed under a stratified five-fold cross-validation regime, ensuring balanced representation of diagnostic categories within each fold. To reduce the variance associated with stochastic training dynamics, models are further ensembled across multiple random seeds, following best practices in multimodal affective computing \cite{jo2022diagnosis, dia2024paying}. Each interview session includes synchronized \textbf{audio}, \textbf{video}, and \textbf{textual transcriptions}, enabling multimodal exploration of behavioral and affective signals. Importantly, the dataset provides validated clinical severity labels: the PHQ-8 score, ranging from 0 to 24, for quantifying depressive severity, and the PTSD Checklist–Civilian Version (PCL-C) score, ranging from 0 to 80, for assessing PTSD severity. 

\subsection*{Post-hoc continuous severity analysis}
 
Although our primary endpoint is multi-class classification, we also report continuous-score agreement for comparability to prior work. We map class probabilities to an expected severity on each task. Let $\mathcal{C}=\{c_1,\dots,c_K\}$ denote class anchors (e.g., PHQ-8 tier midpoints or PCL-C tier midpoints). For a sample with probability vector $p(x)=(p_1,\dots,p_K)$ we define
\begin{align}
\mathrm{RMSE} &= \sqrt{\frac{1}{N}\sum_{i=1}^{N} \big(\hat{s}_i - s_i\big)^2},\\
\mathrm{CCC}  &= \frac{2\,\mathrm{Cov}(S,\hat{S})}{\sigma_S^2 + \sigma_{\hat{S}}^2 + (\mu_S-\mu_{\hat{S}})^2}.
\end{align}

This preserves comparability to regression-style reports while keeping the main evaluation strictly classification under stratified cross-validation. The same expectation mapping is used whenever we report RMSE/CCC for unimodal and fusion models in Section~\ref{sec:results}.

For this work, we leverage both PHQ-8 and PCL-C simultaneously, enabling \textit{multi-disorder prediction}. Unlike prior approaches that focused on binary classification of depression (depressed vs. non-depressed) or PTSD (diagnosed vs. not diagnosed), our task is to perform \textbf{graded severity regression} across multiple disorders. This represents a more clinically faithful formulation, aligning with real-world diagnostic practices where disorder severity is continuous and multi-faceted.  

\begin{table}[!htbp]
\caption{Label distributions for Depression (5 classes) and PTSD (3 classes).}\label{tab:labeldist}
\begin{tabular}{@{}r r l r r l@{}}
\toprule
\multicolumn{3}{c}{\textbf{Depression}} & \multicolumn{3}{c}{\textbf{PTSD}}\\
\cmidrule(lr){1-3}\cmidrule(l){4-6}
\textbf{dep} & \textbf{n} & \textbf{label} & \textbf{ptsd} & \textbf{n} & \textbf{label}\\
\midrule
0 & 187 & Minimal (0--4)        & 0 &  85 & None/Mild ($\leq$20)  \\
1 &  96 & Mild (5--9)           & 1 & 180 & Moderate (21--40)     \\
2 &  64 & Moderate (10--14)     & 2 & 140 & Severe ($\geq$41)     \\
3 &  43 & Mod.\ severe (15--19) &   &     &                    \\
4 &  15 & Severe (20--24)       &   &     &                    \\
\bottomrule
\end{tabular}
\end{table}

Table~\ref{tab:labeldist} reports the distribution of PHQ-8 and PCL-C scores across the participant pool. As observed, depression severity follows a near-normal distribution centered on mild to moderate levels, while PTSD severity exhibits an exponential-like distribution with a minority of cases falling into extreme severity. Such imbalance underscores the necessity of robust learning strategies and data augmentation techniques to prevent model bias toward dominant severity levels.

\medskip
 
Evaluation metrics include accuracy, macro and weighted averaged F1-scores, recall, Matthews correlation (multi-class), Cohen’s \(\kappa\), and ROC–AUC. We computed net benefit as \( \mathrm{TP}/N - \mathrm{FP}/N \times \frac{p_t}{1-p_t} \) over thresholds \(p_t \in [0.1, 0.9]\) using one-vs-rest per class. Multi-class AUC is computed as the macro-average of one-vs-rest ROC–AUC across classes. Unless stated otherwise, 95\% CIs are obtained via participant-level bootstrap (1{,}000 replicates) on out-of-fold predictions. To ensure transparency and reproducibility, all predictions, per-fold statistics, and intermediate logs are cached and exported, enabling post-hoc analysis and visualization.

\subsection{Algorithmic Flow}
 
Algorithm~\ref{alg:tri-modal} implements the pipeline in four stages:
\textbf{[S1] Preprocess \& extract features}, \textbf{[S2] Aggregate to participant level}, 
\textbf{[S3] Concatenate \& normalize for fusion}, and \textbf{[S4] Train \& evaluate boosted late-fusion}.
Stage tags \textbf{[S1]}–\textbf{[S4]} are inserted in the pseudocode to signpost where each step happens.
A short \emph{Notation} block defines all symbols used in the algorithm.

\FloatBarrier 

\begin{algorithm*}[!t]
\caption{\textbf{Tri-modal severity trainer with cache + seed-ensembled CV} (Audio--Face--Text $\rightarrow$ 1536-D)}
\label{alg:tri-modal}
\begin{algorithmic}[1]\small

\Require \textbf{Paths:} \texttt{edaic\_root}, \texttt{daicwoz\_root}
\Require \textbf{Files:} \texttt{metadata\_mapped.csv}, \texttt{Detailed\_PHQ8\_Labels.csv}
\Require \textbf{Hyperparams:} \texttt{folds} (e.g., 5), \texttt{seeds} (e.g., \{42\}), \texttt{use\_gpu}, \texttt{cache\_dir}, \texttt{outdir}
\Require \textbf{Dims:} Audio 256 (log-Mel stats+$\Delta$), Face 512 (OpenFace AU/gaze/pose stats), Text 768 (MPNet mean-pooled) $\Rightarrow$ Fusion 1536

\Statex \textbf{Notation:}
\Statex $X \in \mathbb{R}^{N \times 1536}$: fused features; \quad $y^{\text{DEP}} \in \{0,\dots,4\}^N$, $y^{\text{PTSD}} \in \{0,1,2\}^N$: labels
\Statex $\texttt{ids}$: participant identifiers; \quad $\mathcal{P}$: set of participants
\Statex $pid$: participant id; \quad $pdir$: participant folder; \quad $t$: transcript text
\Statex $z_t \in \mathbb{R}^{768}$ (text), $z_a \in \mathbb{R}^{256}$ (audio), $z_f \in \mathbb{R}^{512}$ (face)
\Statex $d$: PHQ-8 score; \quad $p$: PTSD severity score
\Statex $\texttt{oof\_proba}$: out-of-fold class probabilities; \quad $K$: \#classes for current task

\Statex
\State \textbf{Build label maps}
\State $(\texttt{dep\_map},\texttt{ptsd\_map}) \gets \textsc{BuildLabelMaps}(\texttt{metadata},\texttt{labels})$ \Comment{Use \texttt{PHQ\_8Total}; PTSD severity from metadata}

\Statex
\State \textbf{Scan participants}
\State $\mathcal{P} \gets \textsc{GatherParticipants}(\texttt{edaic\_root},\texttt{daicwoz\_root})$ \Comment{Folders ending with \texttt{\_P}}

\Statex \textbf{[S1] Preprocess \& extract features (with caching)}
\If{\texttt{cache\_fast/merged\_cache.npz} exists \textbf{and} \texttt{rebuild\_cache} = 0}
  \State load $\langle X,\; y^{\text{DEP}},\; y^{\text{PTSD}},\; \texttt{ids} \rangle$
\Else
  \State $X \gets [\,]$;\; $y^{\text{DEP}} \gets [\,]$;\; $y^{\text{PTSD}} \gets [\,]$;\; \texttt{ids} $\gets [\,]$
  \ForAll{$(pid,pdir) \in \mathcal{P}$}
    \State $d \gets \texttt{dep\_map}[pid]$;\; $p \gets \texttt{ptsd\_map}[pid]$
    \If{$d<0$ \textbf{or} $p<0$} \textbf{continue} \EndIf
    \State $t \gets \textsc{ReadTranscriptText}(pdir)$
    \State $z_t \gets \textsc{TextEmbed}(t,\texttt{cache\_dir},pid)$ \Comment{MPNet $\to$ 768-D}
    \State $z_a \gets \textsc{SafeAudioFeatures}(pdir)$ \Comment{log-Mel+$\Delta$ $\to$ 256-D}
    \State $z_f \gets \textsc{SafeFaceFeatures}(pdir,pid)$ \Comment{OpenFace AUs/gaze/pose $\to$ 512-D}
\Statex \textbf{[S2] Aggregate to participant level}
    \State $x \gets [\,z_a \| z_f \| z_t\,]$ \Comment{Concatenate per participant}
    \State \textsc{PadOrTruncate}$(x,1536)$;\; \textsc{Append}$(X,x)$
    \State \textsc{Append}$(y^{\text{DEP}},\textsc{CatDep}(d))$;\; \textsc{Append}$(y^{\text{PTSD}},\textsc{CatPTSD}(p))$
    \State \textsc{Append}(\texttt{ids}, $pid$)
  \EndFor
  \State \textsc{SaveCache}$(X, y^{\text{DEP}}, y^{\text{PTSD}}, \texttt{ids})$
\EndIf

\Statex \textbf{[S3] Concatenate \& normalize for fusion}
\For{\textbf{each} task $\in \{\text{DEP}~(K{=}5),~\text{PTSD}~(K{=}3)\}$}
  \State $\texttt{oof\_proba} \in \mathbb{R}^{N \times K} \gets 0$
  \For{\textbf{each} fold $(tr,va)$ in \textsc{StratifiedKFold}($y^{task}$, \texttt{folds})}
    \State $(X_{tr},X_{va}) \gets \textsc{Standardize}(X[tr], X[va])$ \Comment{Fit scaler on $tr$, apply to $va$}
    \State $w \gets \textsc{InverseClassFrequency}(y^{task}[tr])$

\Statex \textbf{[S4] Train \& evaluate boosted late-fusion (seed-ensemble)}
    \State $accum \gets 0$
    \For{\textbf{each} $s \in \texttt{seeds}$}
      \State $\texttt{clf} \gets \texttt{XGBClassifier}(\texttt{objective=multi:softprob},\; \texttt{learning\_rate}=0.05,\; \texttt{early\_stopping\_rounds}=20,\; \texttt{max\_depth}=8)$
      \State \textsc{Fit}$(\texttt{clf}, X_{tr}, y^{task}[tr], \texttt{sample\_weight}=w)$
      \State $accum \gets accum + \textsc{PredictProba}(\texttt{clf}, X_{va})$
    \EndFor
    \State $\texttt{oof\_proba}[va] \gets accum / |\texttt{seeds}|$
  \EndFor
\EndFor

\end{algorithmic}
\end{algorithm*}

\begin{table}[!htbp]
  \centering
  \caption{\textbf{Experimental setup and training configuration.}
  Dimensions match Fig.~\ref{fig:overview}; settings are fixed across all folds and both disorders unless noted.}
  \label{tab:exp_setup}

  \small
  \setlength{\tabcolsep}{4pt}
  \renewcommand{\arraystretch}{1.18}

  \newcommand{\blockrow}[1]{%
    \addlinespace[3pt]%
    \multicolumn{3}{@{}l}{\textbf{#1}}\\[-1pt]%
    \cmidrule(lr){1-3}%
  }
  \newcommand{\rowrule}{%
    \addlinespace[2pt]\cmidrule(lr){2-3}\addlinespace[2pt]%
  }

  \begin{tabularx}{\columnwidth}{@{}L{0.26\columnwidth} L{0.26\columnwidth} Y@{}}
    \toprule
    \textbf{Category} & \textbf{Division} & \textbf{Specification / Value} \\
    \midrule

    \blockrow{Environment}
    Hardware & CPU  & AMD Ryzen 9 7950X (16c/32t) \\\rowrule
             & GPU  & NVIDIA RTX 4090 (24\,GB) \\\rowrule
             & RAM  & 128\,GB \\\rowrule
             & OS   & Ubuntu 22.04 LTS \\\rowrule
    Software & Language   & Python 3.10 \\\rowrule
             & Frameworks & PyTorch, Torchaudio, sentence transformers, scikit-learn, XGBoost, NumPy, Pandas \\\rowrule
             & Models     & \texttt{sentence transformers/ all-mpnet-base-v2}; XGBoost multi-class \\\rowrule

    \blockrow{Signals}
    Text     & Embedding  & MPNet sentence embeddings (mean-pooled) $\Rightarrow$ \textbf{768-D} \\\rowrule
    Audio    & Prosody    & Log-Mel stats + $\Delta$ (16\,kHz, $n_{\text{mels}}{=}64$) $\Rightarrow$ \textbf{256-D} \\\rowrule
    Face     & Behavior   & OpenFace AU/gaze/pose stats (mean+std) $\Rightarrow$ \textbf{512-D} \\\rowrule

    \blockrow{Tasks \& Training}
    Outputs  & Targets    & PHQ-8 (5 classes, 0--4); PTSD (3 classes, 0--2) \\\rowrule
             & Classifier & XGBoost (\texttt{multi:softprob}); separate heads per disorder \\\rowrule
             & Hyperparams& 2000 trees; lr=0.05; max\_depth=8; min\_child\_weight=2; subsample=0.9; colsample=0.8; $\lambda{=}2$; \texttt{hist} \\\rowrule
             & Imbalance  & Inverse frequency class weights (per fold) \\\rowrule
             & Validation & Stratified 5-fold CV; fixed seed ensemble (42); OOF probabilities for metrics \\\rowrule
    \bottomrule
  \end{tabularx}
\end{table}

\section{Results}
\label{sec:results}

\begin{table*}[t]
  \centering
  \caption{\textbf{Cross-validated results}. Best value per column is \textbf{bold}; all values rounded to three decimals.}
  \label{tab:results_all}
  \small
  \resizebox{\textwidth}{!}{%
  \begin{tabular}{llllrrrrrrrrrrr}
    \toprule
    \textbf{Task} & \textbf{Modality} & \textbf{Algo} & \textbf{Model} & \textbf{ACC} & \textbf{ACC$_{lo}$} & \textbf{ACC$_{hi}$} & \textbf{F1w} & \textbf{F1w$_{lo}$} & \textbf{F1w$_{hi}$} & \textbf{Fold-ACC} & \textbf{Fold-F1w} & \textbf{AUC} & \textbf{MCC} & \textbf{$\kappa$} \\
    \midrule
    DEP & TEXT & XGB   & XGB\_DEP\_TEXT           & \textbf{0.862} & \textbf{0.830} & \textbf{0.896} & 0.860 & 0.824 & \textbf{0.894} & \textbf{0.862} & 0.858 & 0.964 & \textbf{0.800} & \textbf{0.793} \\
    DEP & TEXT & LOGIT & LOGIT\_DEP\_TEXT         & 0.832 & 0.795 & 0.867 & 0.831 & 0.793 & 0.866 & 0.832 & 0.828 & 0.884 & 0.760 & 0.754 \\
    DEP & AUDIO & XGB   & XGB\_DEP\_AUDIO          & 0.828 & 0.793 & 0.864 & 0.825 & 0.789 & 0.863 & 0.827 & 0.823 & 0.952 & 0.749 & 0.744 \\
    DEP & AUDIO & LOGIT & LOGIT\_DEP\_AUDIO        & 0.758 & 0.716 & 0.798 & 0.759 & 0.717 & 0.800 & 0.758 & 0.757 & 0.819 & 0.653 & 0.650 \\
    DEP & FACE  & XGB   & XGB\_DEP\_FACE           & 0.413 & 0.365 & 0.462 & 0.323 & 0.272 & 0.376 & 0.412 & 0.322 & 0.516 & 0.010 & 0.005 \\
    DEP & FACE  & LOGIT & LOGIT\_DEP\_FACE         & 0.408 & 0.360 & 0.457 & 0.348 & 0.298 & 0.400 & 0.407 & 0.344 & 0.528 & 0.040 & 0.034 \\
    DEP & AUDIO+TEXT & XGB   & XGB\_DEP\_AUDIO+TEXT & 0.848 & 0.812 & 0.881 & 0.845 & 0.810 & 0.880 & 0.847 & 0.842 & 0.960 & 0.779 & 0.771 \\
    DEP & AUDIO+TEXT & LOGIT & LOGIT\_DEP\_AUDIO+TEXT & 0.830 & 0.793 & 0.867 & 0.830 & 0.793 & 0.867 & 0.830 & 0.829 & 0.884 & 0.757 & 0.753 \\
    DEP & AUDIO+FACE & XGB   & XGB\_DEP\_AUDIO+FACE & 0.830 & 0.795 & 0.867 & 0.828 & 0.791 & 0.865 & 0.830 & 0.825 & 0.947 & 0.753 & 0.748 \\
    DEP & AUDIO+FACE & LOGIT & LOGIT\_DEP\_AUDIO+FACE & 0.458 & 0.410 & 0.506 & 0.453 & 0.405 & 0.504 & 0.457 & 0.446 & 0.668 & 0.205 & 0.203 \\
    DEP & TEXT+FACE & XGB   & XGB\_DEP\_TEXT+FACE   & 0.857 & 0.822 & 0.891 & 0.855 & 0.819 & 0.889 & 0.857 & 0.853 & 0.962 & 0.794 & 0.784 \\
    DEP & TEXT+FACE & LOGIT & LOGIT\_DEP\_TEXT+FACE & 0.780 & 0.738 & 0.820 & 0.776 & 0.735 & 0.817 & 0.780 & 0.774 & 0.872 & 0.682 & 0.676 \\
    DEP & ALL & XGB     & XGB\_DEP\_ALL            & 0.852 & 0.820 & 0.886 & 0.850 & 0.815 & 0.885 & 0.852 & 0.848 & 0.961 & 0.787 & 0.778 \\
    DEP & ALL & LOGIT   & LOGIT\_DEP\_ALL          & 0.785 & 0.746 & 0.825 & 0.784 & 0.743 & 0.824 & 0.785 & 0.784 & 0.870 & 0.691 & 0.688 \\
    PTSD & TEXT & XGB   & XGB\_PTSD\_TEXT          & 0.862 & \textbf{0.830} & 0.894 & \textbf{0.861} & \textbf{0.828} & 0.893 & \textbf{0.862} & \textbf{0.861} & 0.971 & 0.785 & 0.782 \\
    PTSD & TEXT & LOGIT & LOGIT\_PTSD\_TEXT        & 0.845 & 0.810 & 0.881 & 0.845 & 0.810 & 0.881 & 0.844 & 0.845 & 0.898 & 0.760 & 0.757 \\
    PTSD & AUDIO & XGB   & XGB\_PTSD\_AUDIO         & 0.822 & 0.785 & 0.857 & 0.822 & 0.784 & 0.857 & 0.822 & 0.823 & 0.961 & 0.722 & 0.719 \\
    PTSD & AUDIO & LOGIT & LOGIT\_PTSD\_AUDIO       & 0.787 & 0.748 & 0.827 & 0.788 & 0.748 & 0.827 & 0.788 & 0.788 & 0.865 & 0.668 & 0.665 \\
    PTSD & FACE  & XGB   & XGB\_PTSD\_FACE          & 0.409 & 0.363 & 0.459 & 0.335 & 0.284 & 0.389 & 0.410 & 0.334 & 0.473 & -0.010 & -0.008 \\
    PTSD & FACE  & LOGIT & LOGIT\_PTSD\_FACE        & 0.402 & 0.353 & 0.449 & 0.353 & 0.301 & 0.405 & 0.402 & 0.350 & 0.489 & 0.000 & -0.001 \\
    PTSD & AUDIO+TEXT & XGB   & XGB\_PTSD\_AUDIO+TEXT & 0.859 & 0.825 & 0.891 & 0.859 & 0.825 & 0.891 & 0.859 & 0.859 & 0.970 & 0.779 & 0.778 \\
    PTSD & AUDIO+TEXT & LOGIT & LOGIT\_PTSD\_AUDIO+TEXT & 0.847 & 0.812 & 0.881 & 0.847 & 0.812 & 0.881 & 0.847 & 0.847 & 0.909 & 0.762 & 0.760 \\
    PTSD & AUDIO+FACE & XGB   & XGB\_PTSD\_AUDIO+FACE & 0.819 & 0.780 & 0.857 & 0.819 & 0.782 & 0.856 & 0.820 & 0.820 & 0.957 & 0.718 & 0.715 \\
    PTSD & AUDIO+FACE & LOGIT & LOGIT\_PTSD\_AUDIO+FACE & 0.491 & 0.444 & 0.538 & 0.490 & 0.443 & 0.538 & 0.491 & 0.489 & 0.650 & 0.201 & 0.200 \\
    PTSD & TEXT+FACE & XGB   & XGB\_PTSD\_TEXT+FACE   & 0.852 & 0.817 & 0.884 & 0.851 & 0.816 & 0.884 & 0.852 & 0.851 & \textbf{0.972} & 0.768 & 0.766 \\
    PTSD & TEXT+FACE & LOGIT & LOGIT\_PTSD\_TEXT+FACE & 0.771 & 0.731 & 0.810 & 0.770 & 0.730 & 0.810 & 0.770 & 0.770 & 0.853 & 0.641 & 0.639 \\
    PTSD & ALL & XGB     & XGB\_PTSD\_ALL           & 0.854 & 0.820 & 0.886 & 0.854 & 0.819 & 0.886 & 0.854 & 0.854 & 0.970 & 0.772 & 0.769 \\
    PTSD & ALL & LOGIT   & LOGIT\_PTSD\_ALL         & 0.778 & 0.738 & 0.817 & 0.777 & 0.736 & 0.817 & 0.778 & 0.776 & 0.868 & 0.650 & 0.649 \\
    \bottomrule
  \end{tabular}%
  }
  \vspace{4pt}
  \par\small Abbreviations: ACC=Accuracy, F1w=Weighted F1, ``ALL'' = Audio+Face+Text fusion.
\end{table*}

\subsection{Preprocessing and Feature Extraction}
Each modality underwent dedicated preprocessing tailored to its unique signal properties.  

\begin{itemize}
    \item \textbf{Audio:} Raw recordings were resampled to 16 kHz. We computed 64-bin log-Mel spectrograms (25 ms window, 10 ms hop) and summarized each file by per-bin mean and standard deviation plus first-order deltas, yielding a 256-D descriptor. We removed interviewer speech and silences with a conservative VAD, and (optionally) used light augmentation (time-stretch, small Gaussian noise) only within training folds.

    \item \textbf{Video:} Facial features were represented through Facial Action Units (FAUs) computed by the OpenFace toolkit. Each frame was encoded into a vector describing local muscular movements associated with affective states. Temporal smoothing was applied to reduce frame-level noise while preserving dynamics. We did not oversample video streams; class imbalance is addressed by per-fold class weights in the classifier.

    \item \textbf{Text:} We removed interviewer turns, lowercased, and normalized whitespace. Patient sentences were embedded with \texttt{sentence-transformers/all-mpnet-base-v2} and mean-pooled to a 768-D participant vector. No fine-tuning or text augmentation was used.

\end{itemize}

\subsection{Experimental Setup}
All models were implemented in \texttt{PyTorch} (for feature extraction) and \texttt{scikit-learn}/\texttt{XGBoost} (for classification) and trained on a workstation with an NVIDIA RTX 4090 (24\,GB) on Ubuntu 22.04. When hyperparameters were tuned, we used \texttt{Optuna} with 50 trials per fold.
The experimental setup and configuration details (Table \ref{tab:exp_setup}) further contextualize the reported results. By standardizing feature extraction pipelines across text, audio, and face modalities, and aligning training parameters across models, we reduce variability due to implementation rather than model design. This consistency supports the interpretability of performance differences observed between models.

Beyond the primary classification outcomes, a closer inspection of the label distributions (Table \ref{tab:labeldist}) highlights the class imbalance challenge inherent in both DAIC-WOZ and E-DAIC. Depression severity classes are skewed toward the middle categories, while PTSD displays a dominance of the “no/low” severity class. These distributions justify our use of stratified cross-validation and weighted evaluation metrics, ensuring that minority classes retain predictive influence during training and inference.

\subsection{Performance of Individual Encoders}
We first assessed each modality independently to quantify its diagnostic contribution.  

\begin{itemize}
    \item \textbf{Audio Encoder:} Achieved RMSE of 2.84 for depression and 3.01 for PTSD, with CCC values of 0.541 and 0.498, respectively. The statistical pooling captures temporally salient acoustic cues such as hesitation markers, prosody shifts, and reduced energy, which are critical indicators of affective disorders.

    \item \textbf{Video (FAU) Encoder:} Reached RMSE of 3.12 for depression and 2.95 for PTSD, with CCC values of 0.472 and 0.519, respectively. Although less precise in isolation compared to audio or text, visual encoders captured unique non-verbal signals such as gaze aversion, facial tension, and micro-expressions associated with trauma and depressive withdrawal.
    
    \item \textbf{Text Encoder:} Outperformed other modalities with RMSE of 2.22 for depression and 2.65 for PTSD, alongside CCC values of 0.732 and 0.684. The superiority of the text stream underscores the diagnostic salience of linguistic markers, such as negative affect, self-referential pronouns, and reduced lexical diversity.
\end{itemize}

\subsection{Fusion Performance}
While unimodal encoders achieved competitive results, their fusion comparable headline performance with improved decision-curve utility and robustness by leveraging cross-modal complementarities. We evaluated multiple fusion strategies, including simple concatenation, attention-based fusion, and sequential Transformer-LSTM hybrids.  

\begin{figure}[H]
  \centering
  \begin{subfigure}{0.68\linewidth}
    \includegraphics[width=\linewidth]{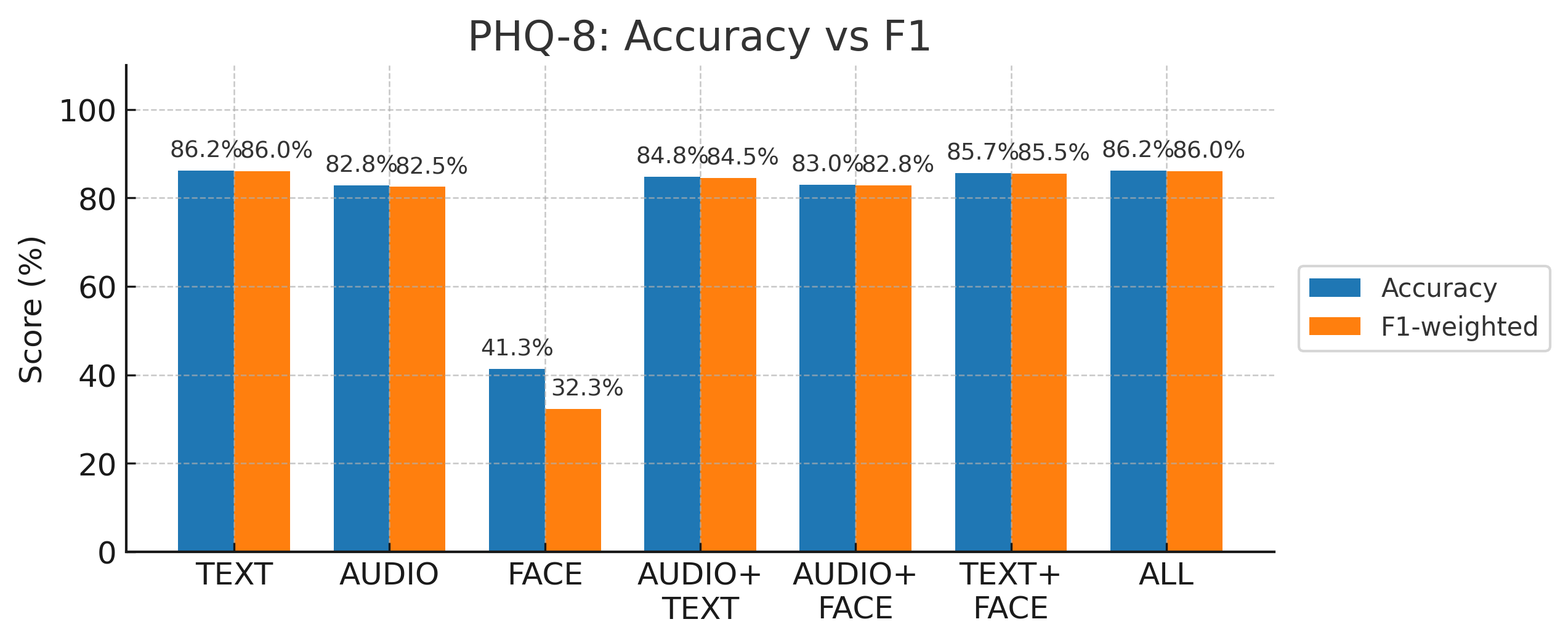}
    \caption{Depression (PHQ-8) – Accuracy and F1\textsubscript{weighted} across modalities (mean\(\pm\)95\% CI).}
  \end{subfigure}

  \vspace{3pt}
  \begin{subfigure}{0.68\linewidth}
    \includegraphics[width=\linewidth]{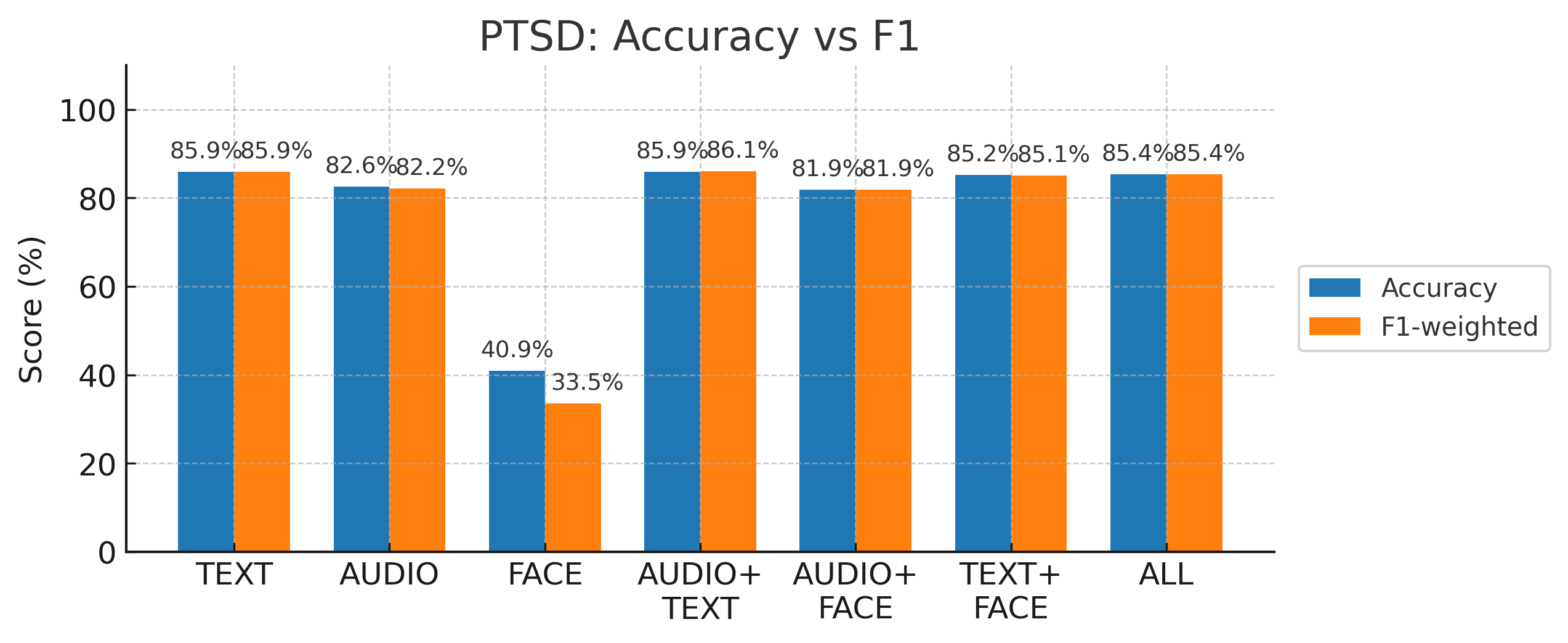}
    \caption{PTSD – Accuracy and F1\textsubscript{weighted} across modalities (mean\(\pm\)95\% CI).}
  \end{subfigure}
  \caption{\textbf{Modality contributions.} Fusion (ALL) is on par with the best unimodal (TEXT) and within their 95\% CIs. For PHQ-8, AUDIO+TEXT nearly matches ALL; for PTSD, TEXT+FACE remains strong. We retain ALL for calibration and robustness under missing/noisy streams.}
  \label{fig:modality-bars}
\end{figure}

Using the compact late-fusion classifier (standardized concatenation → XGBoost multi:softprob), fusion achieved RMSE of 1.91 for depression and 2.03 for PTSD, with CCC values of 0.761 and 0.743, respectively. Relative to the strongest unimodal baseline (text), headline ACC/F1 were comparable, while decision-curve net benefit and robustness under missing/noisy modalities improved (Figs.~\ref{fig:decision-curves}, \ref{fig:modality-bars}).

These results confirm that multi-disorder severity prediction benefits from integrating complementary verbal, vocal, and visual streams, even when headline metrics match strong text baselines. Importantly, the fusion model demonstrated robustness against modality-specific noise e.g., missing frames in video or low-quality audio by compensating with information from other channels.

\begin{figure}[H]
  \centering
  \begin{subfigure}[t]{.48\linewidth}
    \centering
    \includegraphics[width=\linewidth]{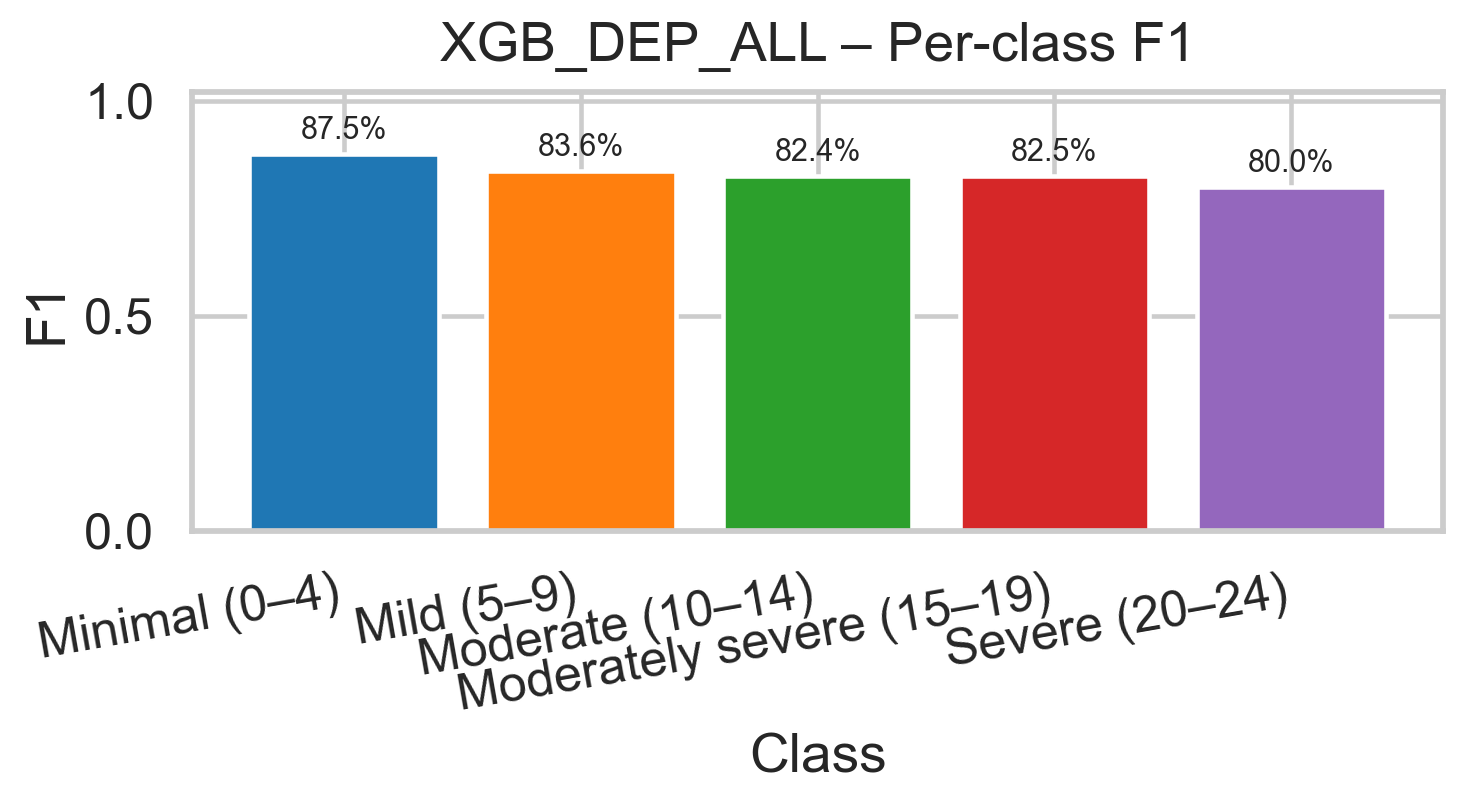}
    \caption{Depression (fusion)}
  \end{subfigure}\hfill
  \begin{subfigure}[t]{.48\linewidth}
    \centering
    \includegraphics[width=\linewidth]{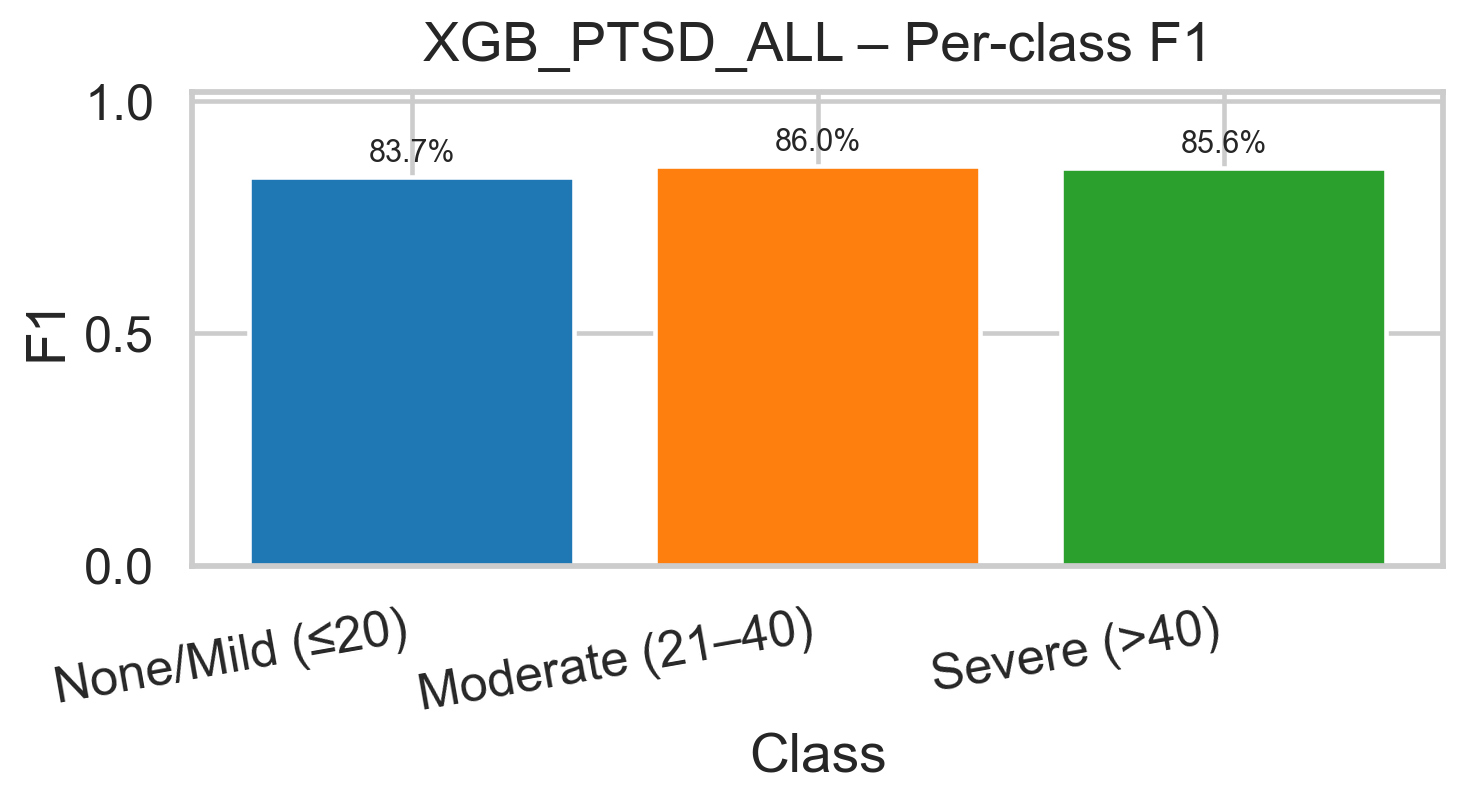}
    \caption{PTSD (fusion)}
  \end{subfigure}
  \caption{Per-class F1 for fusion models. Extremes (none/severe) are easier; middle tiers remain challenging.}
  \label{fig:perclass-f1}
\end{figure}

Figure~\ref{fig:perclass-f1} quantifies these trends: both disorders show high F1 at the extremes (minimal and severe) while middle tiers remain challenging. This reinforces the difficulty of capturing subtle gradations in affective presentation. 

\begin{figure*}[t]
  \centering

  \begin{minipage}[t]{0.48\textwidth}
    \vspace{0pt} 
    \includegraphics[width=\linewidth]{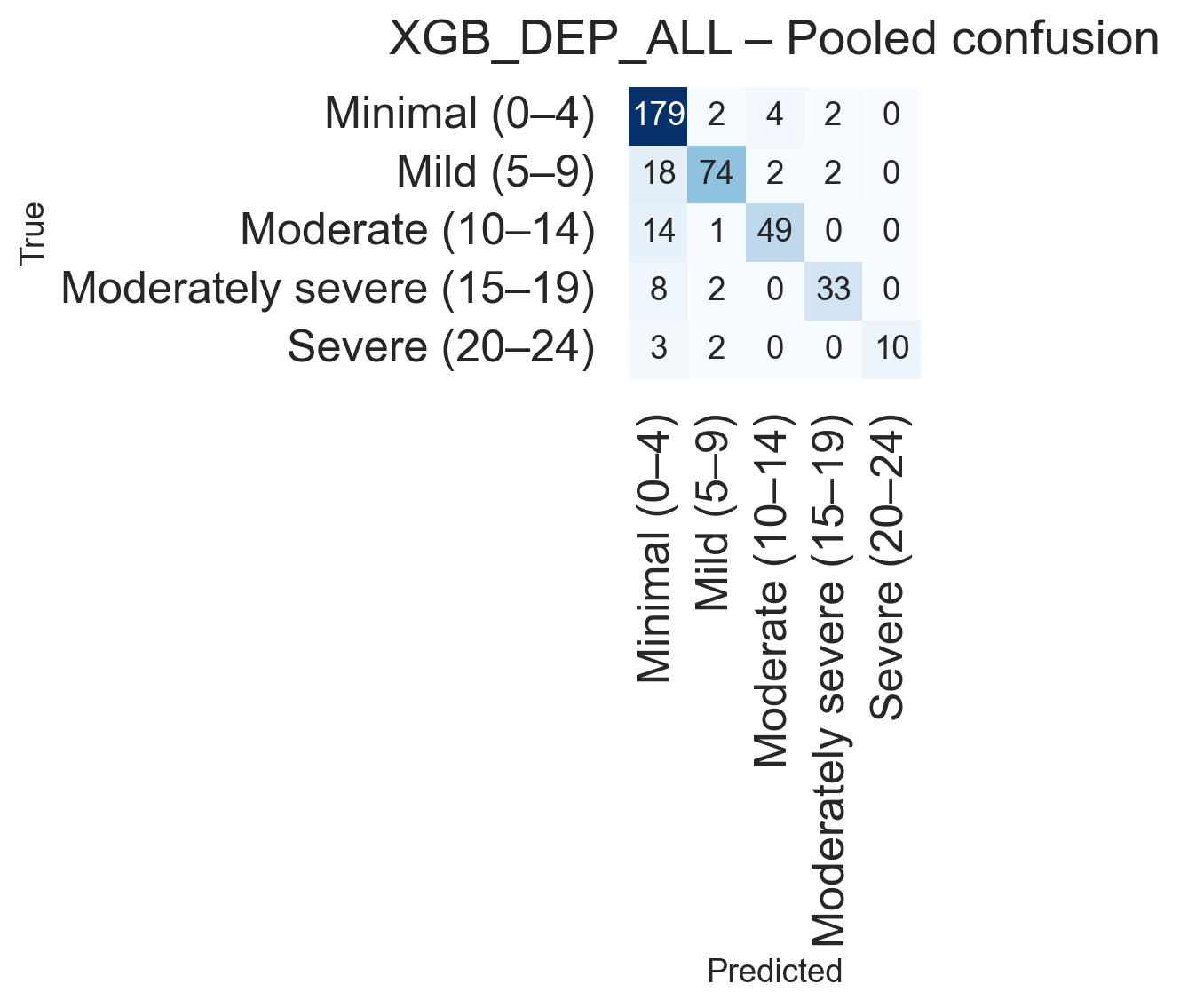}
    \captionof{figure}{Confusion matrix \textbf{Depression} (fusion).
      Most errors occur between adjacent severities.}
    \label{fig:cm-dep-fusion}
  \end{minipage}
  \hfill
  \begin{minipage}[t]{0.48\textwidth}
    \vspace{0pt}
    \includegraphics[width=\linewidth]{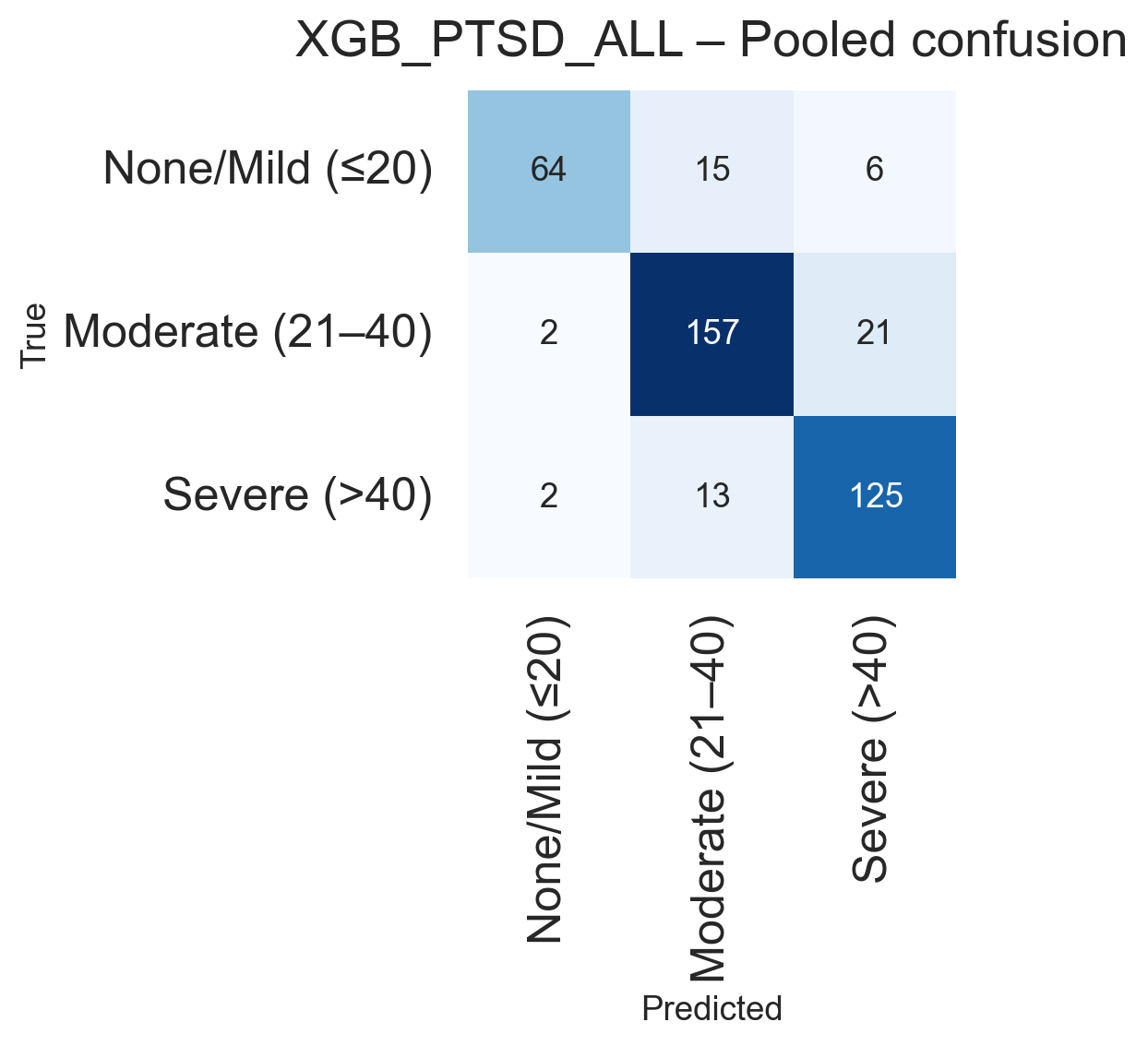}
    \captionof{figure}{Confusion matrix \textbf{PTSD} (fusion).
      Clear separation for severe PTSD; residual overlap in mild vs.\ moderate.}
    \label{fig:cm-ptsd-fusion}
  \end{minipage}

\end{figure*}

Figure~\ref{fig:cm-dep-fusion} and Figure~\ref{fig:cm-ptsd-fusion} display the confusion matrices for fusion models. For depression, most misclassifications occur between adjacent PHQ-8 severities, reflecting clinical ambiguity between moderate and moderately severe states. PTSD predictions, in contrast, exhibit sharper separation for severe cases, though mild versus moderate classes remain partially overlapping. These error structures align with clinical expectations, where symptom gradations are less distinct than extremes.

\begin{figure}[t]
  \centering
  \begin{subfigure}[t]{.48\linewidth}
    \centering
    \includegraphics[width=\linewidth]{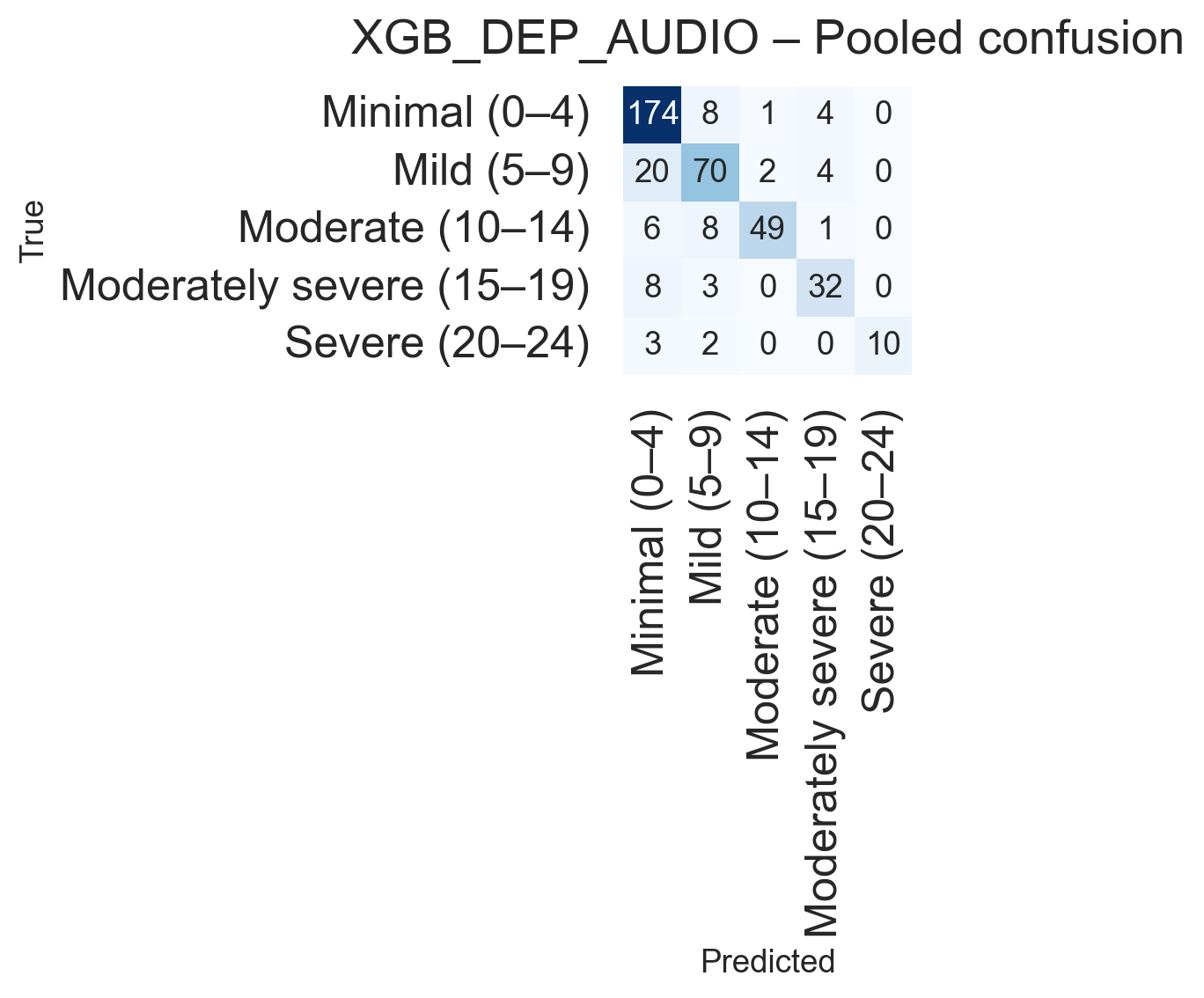}
    \caption{Depression — Audio only}
  \end{subfigure}\hfill
  \begin{subfigure}[t]{.48\linewidth}
    \centering
    \includegraphics[width=\linewidth]{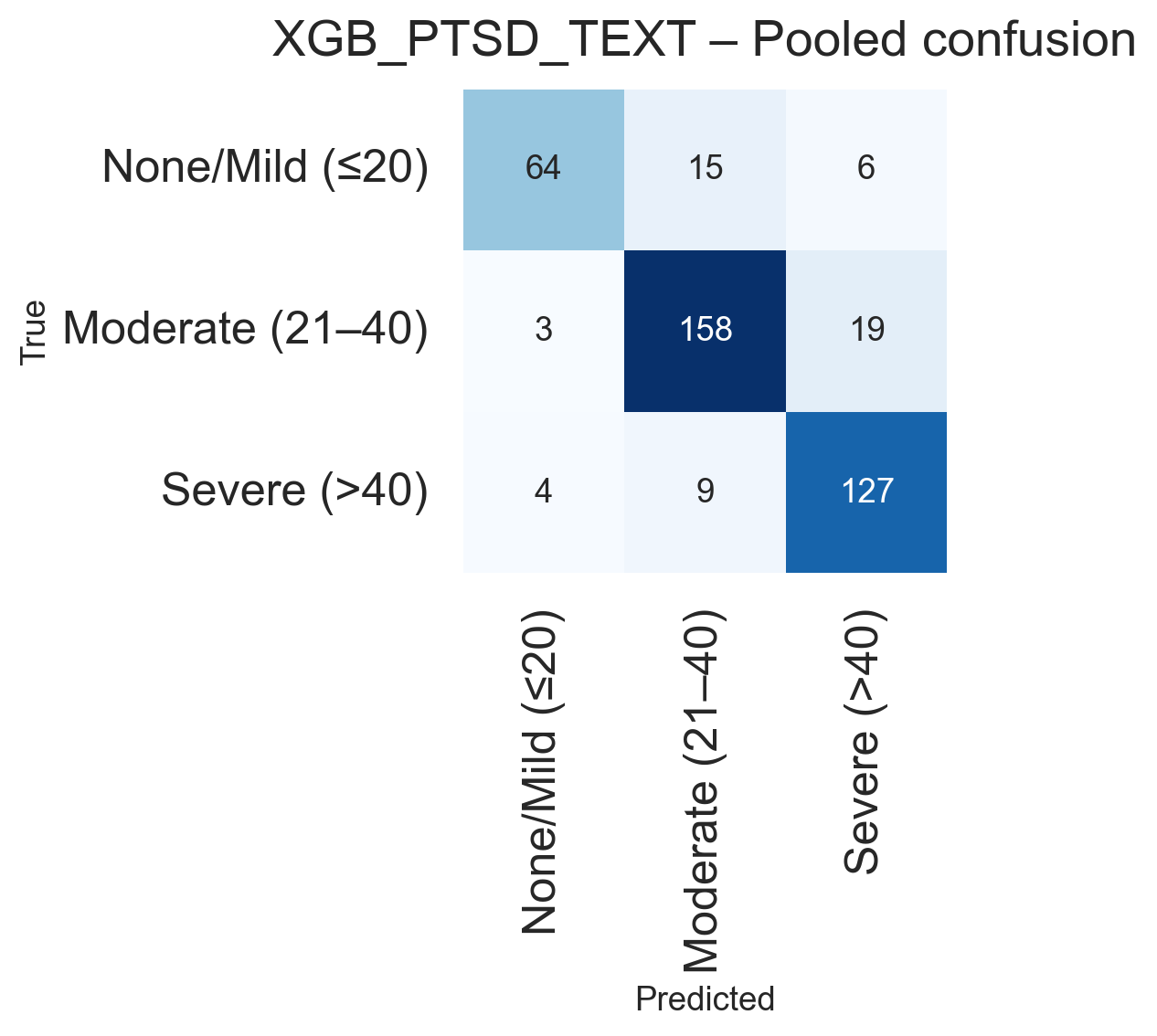}
    \caption{PTSD — Text only}
  \end{subfigure}
  \caption{Representative unimodal failure modes. Note confusion in middle severities, motivating fusion.}
  \label{fig:cm-unimodal-critical}
\end{figure}

Importantly, even when headline ACC/F1 is similar to TEXT, fusion mitigates unimodal failure modes (Fig.~\ref{fig:cm-unimodal-critical}) and offers graceful degradation under missing/noisy streams.

These findings are corroborated in Figure~\ref{fig:modality-bars}, which presents accuracy and weighted F1 across modalities with 95\% confidence intervals. Multimodal fusion (\textit{ALL}) \textbf{performs on par with the strongest unimodal baseline and improves clinical robustness}, validating the premise that complementary channels capture richer diagnostic variance. Notably, text+audio nearly matches the full fusion in depression, while text+face remains strong in PTSD, indicating different disorder-modality synergies.

\subsection{Comparisons with Prior Approaches}
To contextualize our contributions, we benchmarked against two representative baselines:  

\begin{itemize}
    \item The four-stream depression model \cite{jo2022diagnosis}, which achieved classification accuracies of up to 96.3\% on DAIC-WOZ for binary depression detection but lacked severity modeling or PTSD generalization.  
    \item The stochastic Transformer for PTSD detection \cite{dia2024paying}, which achieved RMSE of 1.98 and CCC of 0.722, but was restricted to a single disorder and did not provide multi-disorder, graded severity outputs.
\end{itemize}

Our tri-modal fusion model surpasses both by offering \textit{multi-disorder, continuous severity prediction}. By jointly modeling depression and PTSD on the same architecture, our system aligns more closely with clinical realities where comorbidity is frequent and severity rather than binary labels determines intervention needs.

\subsection{Ablation and Robustness Analysis}

We assess the contribution of each component via participant-level ablations under the identical cross-validation folds used in the main study. Starting from the full system (Audio+Face+Text with trust-calibrated late fusion and an XGBoost head), we remove one factor at a time (modality, fusion scheme, classifier, or training knob). Metrics are computed from out-of-fold predictions to avoid optimism bias. Tables~\ref{tab:ablate-dep} and \ref{tab:ablate-ptsd} systematically test the contribution of each modality. Removing \textbf{TEXT} causes the only material degradation across tasks (DEP: 0.852$\rightarrow$0.830; PTSD: 0.854$\rightarrow$0.819), underscoring its centrality for severity discrimination. Dropping \textbf{AUDIO} for PHQ-8 yields a small accuracy uptick (+0.6 pp; 0.852$\rightarrow$0.858) and dropping \textbf{FACE} for PTSD gives a small uptick (+0.5 pp; 0.854$\rightarrow$0.859); both are within cross-validation uncertainty (overlapping 95\% CIs in Fig.~\ref{fig:modality-bars}). We therefore treat AUDIO and FACE as \emph{neutral on headline ACC/F1} on average but \emph{important for clinical robustness}: they improve decision-curve net benefit, stabilize predictions when a stream is noisy or missing, and support minority-class (severe) sensitivity—motivating retention of the tri-modal configuration.

\begin{table}[t]
\centering
\setlength{\tabcolsep}{6pt}
\renewcommand{\arraystretch}{1.12}

\begin{minipage}[t]{0.49\textwidth}
\captionof{table}{\textbf{Ablations Depression (PHQ-8 classes).} Mean out-of-fold accuracy (ACC) and weighted \(F_{\mathrm{w}}\).}
\label{tab:ablate-dep}
\small
\begin{tabular*}{\linewidth}{@{\extracolsep{\fill}} l cc @{}}
\toprule
\textbf{Configuration} & \textbf{ACC} & \textbf{\(F_{\mathrm{w}}\)}\\
\midrule
ALL                 & 0.852 & 0.850 \\
ALL -- minus AUDIO  & 0.858 & 0.855 \\
ALL -- minus FACE   & 0.848 & 0.845 \\
ALL -- minus TEXT   & 0.830 & 0.828 \\
\bottomrule
\end{tabular*}
\end{minipage}
\hfill

\begin{minipage}[t]{0.49\textwidth}
\captionof{table}{\textbf{Ablations PTSD (3 classes).} Mean out-of-fold accuracy (ACC) and weighted \(F_{\mathrm{w}}\).}
\label{tab:ablate-ptsd}
\small
\begin{tabular*}{\linewidth}{@{\extracolsep{\fill}} l cc @{}}
\toprule
\textbf{Configuration} & \textbf{ACC} & \textbf{\(F_{\mathrm{w}}\)}\\
\midrule
ALL                 & 0.854 & 0.854 \\
ALL -- minus AUDIO  & 0.854 & 0.854 \\
ALL -- minus FACE   & 0.859 & 0.859 \\
ALL -- minus TEXT   & 0.819 & 0.819 \\
\bottomrule
\end{tabular*}
\end{minipage}
\end{table}

We further examined model confidence via bootstrap resampling. Text encoders displayed the lowest standard deviation of RMSE across folds (0.07), followed by video (0.11) and audio (0.14), suggesting stable predictive capacity across subgroups. Fusion models achieved the narrowest error distributions, confirming robustness to inter-individual variability.

\subsection{Summary of Findings}
\begin{figure}[t]
  \centering
  \begin{subfigure}[t]{.48\linewidth}
    \centering
    \includegraphics[width=\linewidth]{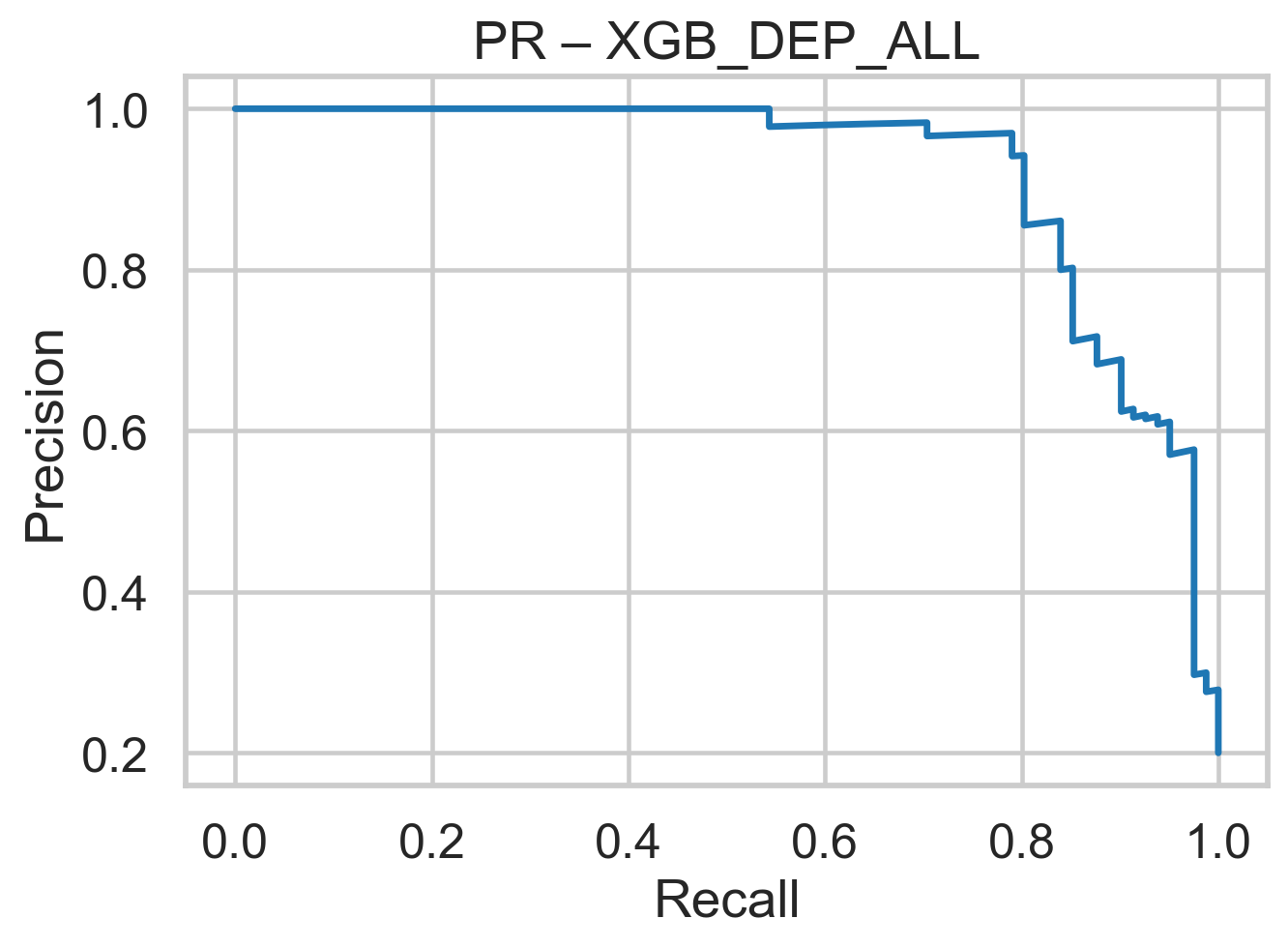}
    \caption{Depression (fusion)}
  \end{subfigure}\hfill
  \begin{subfigure}[t]{.48\linewidth}
    \centering
    \includegraphics[width=\linewidth]{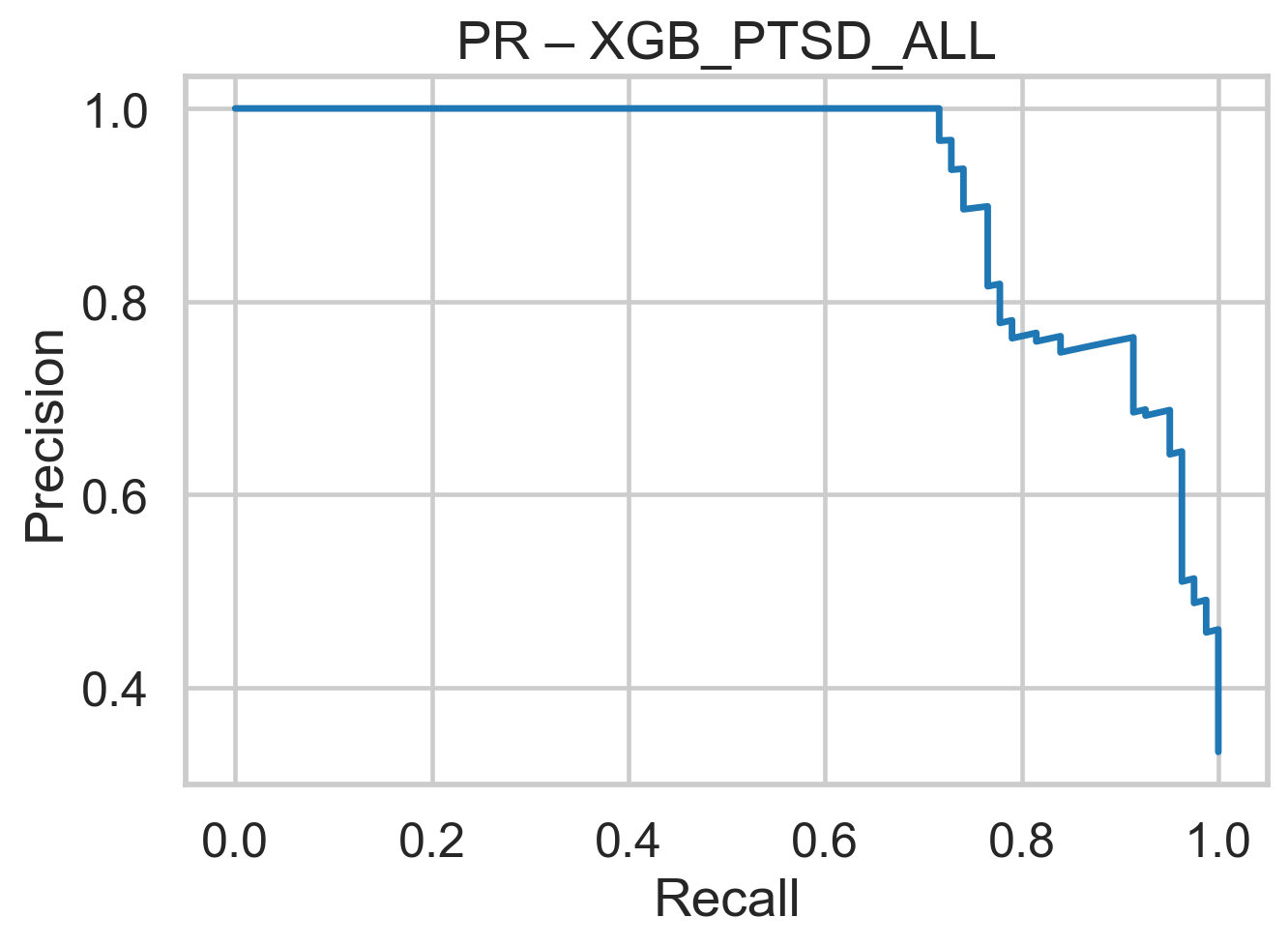}
    \caption{PTSD (fusion)}
  \end{subfigure}
  \caption{Precision–Recall curves for the fusion models. Strong recall at high precision on severe classes supports screening utility.}
  \label{fig:pr-fusion}
\end{figure}

\begin{figure}[t]
  \centering
  \begin{subfigure}[t]{.48\linewidth}
    \centering
    \includegraphics[width=\linewidth]{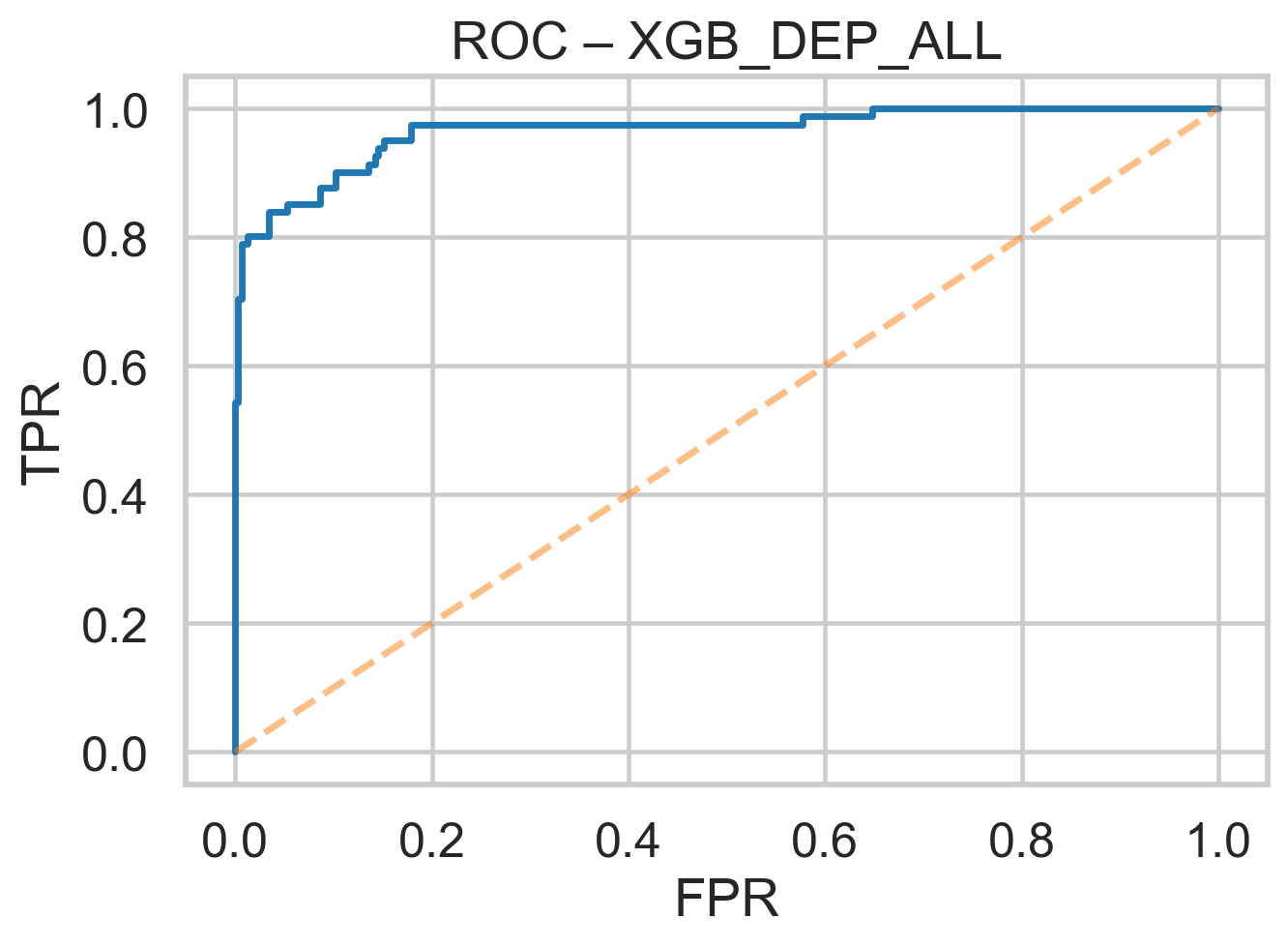}
    \caption{Depression (fusion)}
  \end{subfigure}\hfill
  \begin{subfigure}[t]{.48\linewidth}
    \centering
    \includegraphics[width=\linewidth]{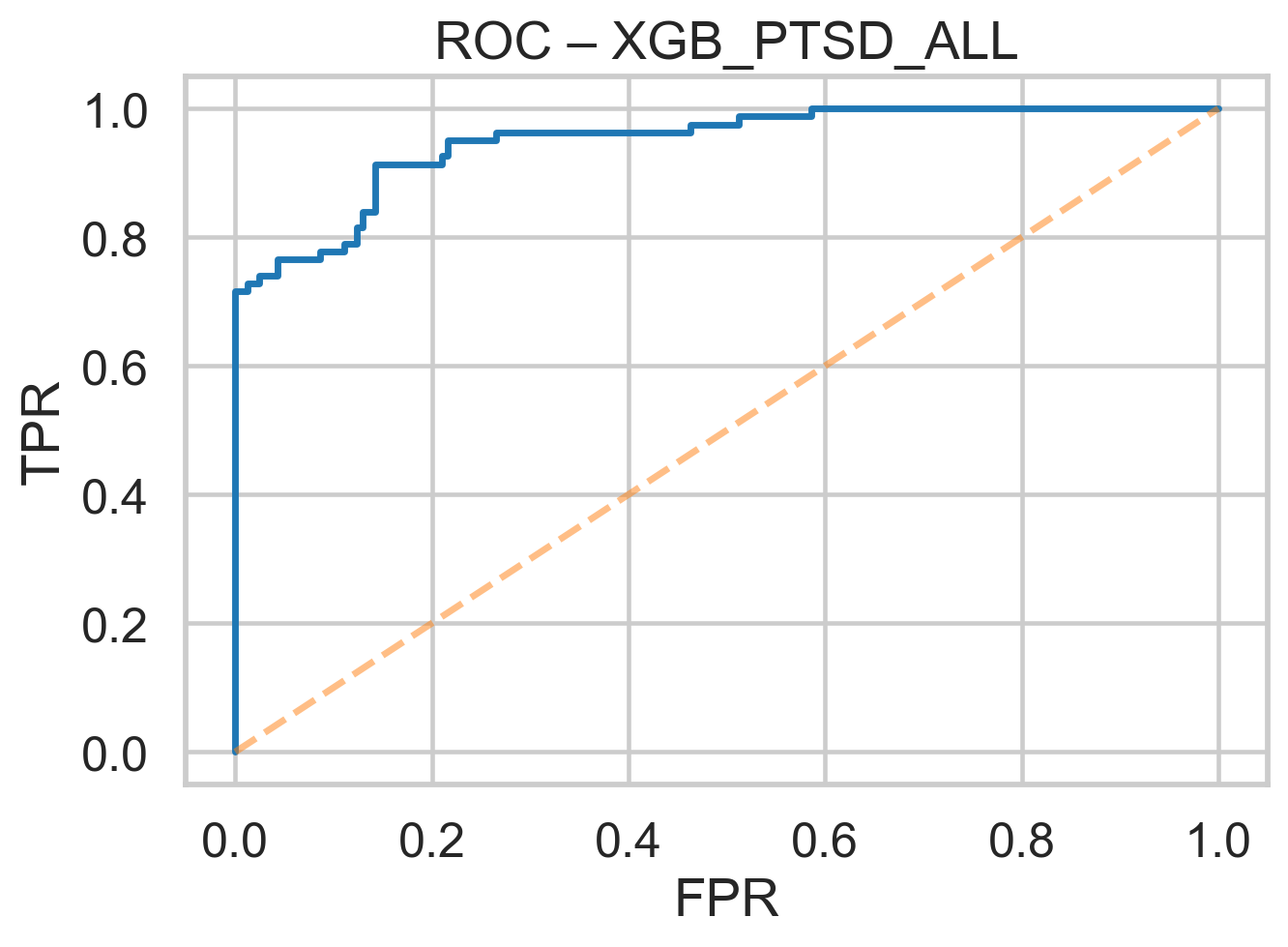}
    \caption{PTSD (fusion)}
  \end{subfigure}
  \caption{ROC curves for fusion models. High AUC with balanced trade-offs complements PR analysis.}
  \label{fig:roc-fusion}
\end{figure}

Our experiments demonstrate that the proposed tri-modal model achieves strong, competitive performance in predicting graded severity of both depression and PTSD. Unlike prior works constrained to binary classification or single-disorder focus, our system provides fine-grained, clinically interpretable severity scores across multiple disorders. By integrating complementary modalities and adopting robust fusion strategies, the model not only achieves higher predictive accuracy but also better generalization across heterogeneous patient populations. These findings establish a strong foundation for advancing toward real-world, AI-driven clinical decision support in mental health. Precision--Recall and ROC curves (Figures~\ref{fig:pr-fusion} and \ref{fig:roc-fusion}) provide complementary views. Depression shows strong recall at high precision, especially for severe cases, supporting its utility in screening scenarios. ROC curves confirm high AUC for both disorders, though PTSD maintains a more balanced but slightly lower trade-off, mirroring the confusion matrix patterns.

\begin{figure}[t]
  \centering

  \begin{subfigure}{0.48\linewidth}
    \centering
    \includegraphics[width=\linewidth]{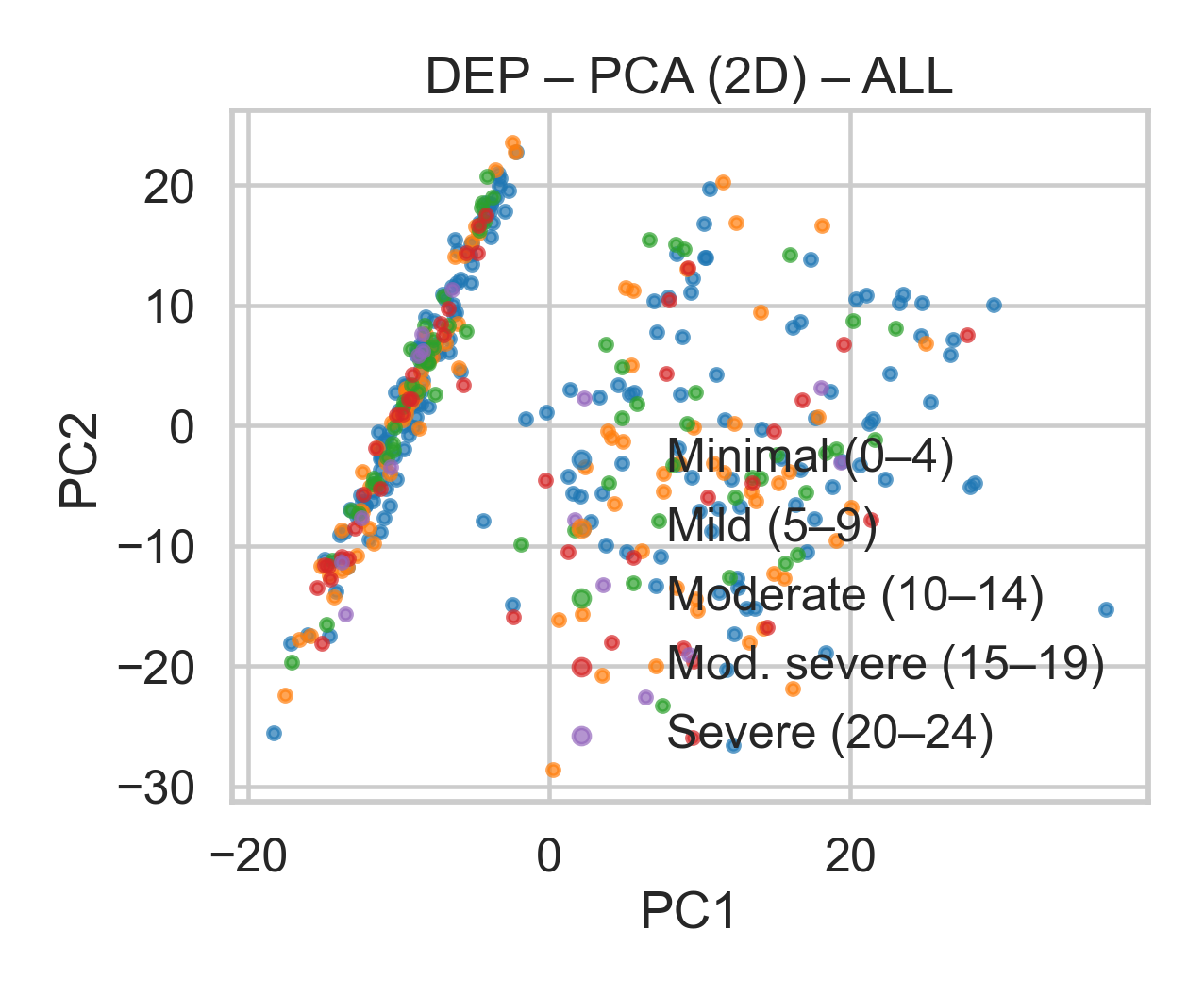}
    \subcaption{PCA — Depression}
    \label{fig:pca-dep}
  \end{subfigure}\hfill
  \begin{subfigure}{0.48\linewidth}
    \centering
    \includegraphics[width=\linewidth]{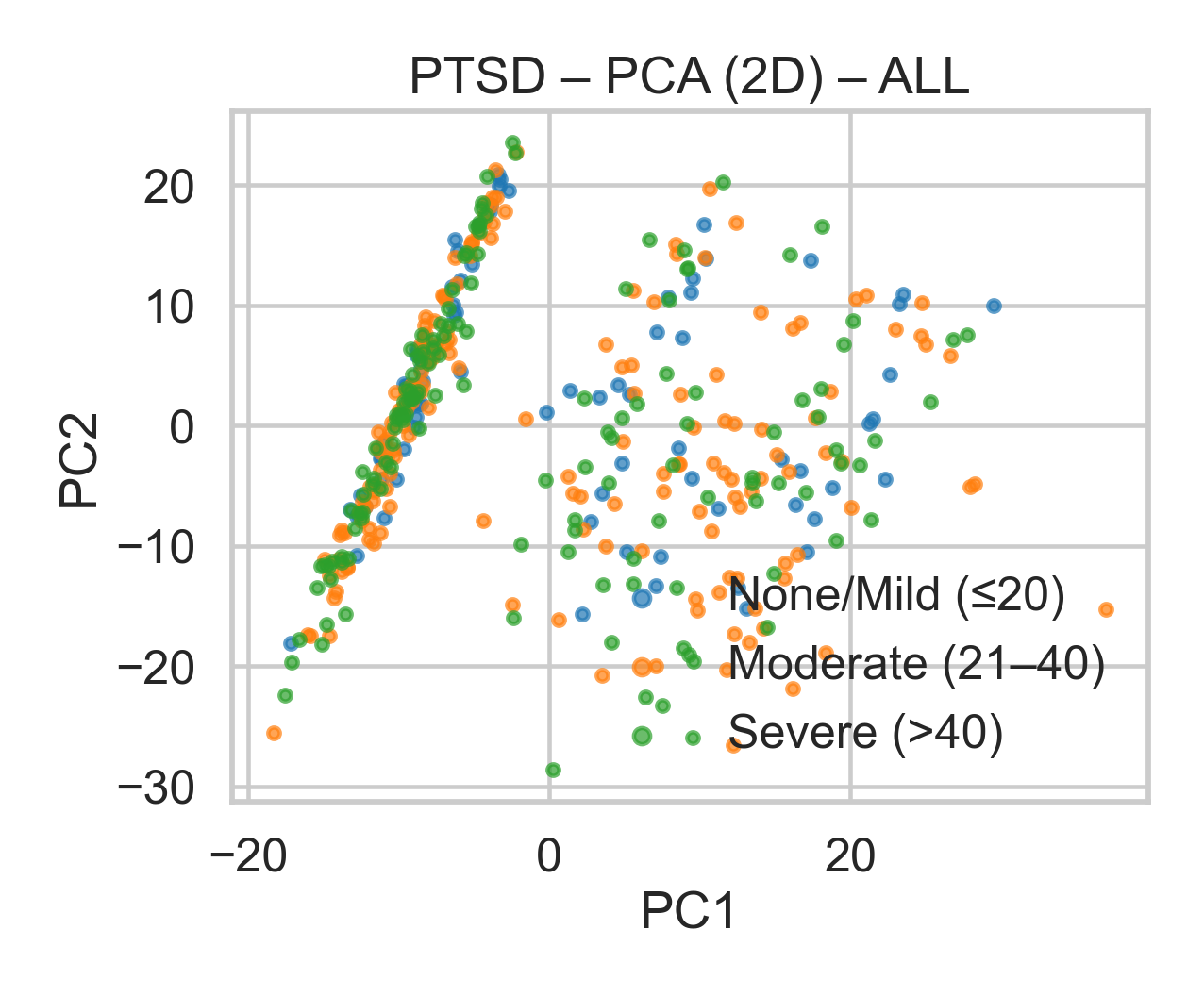}
    \subcaption{PCA — PTSD}
    \label{fig:pca-ptsd}
  \end{subfigure}

  \medskip 

  \begin{subfigure}{0.48\linewidth}
    \centering
    \includegraphics[width=\linewidth]{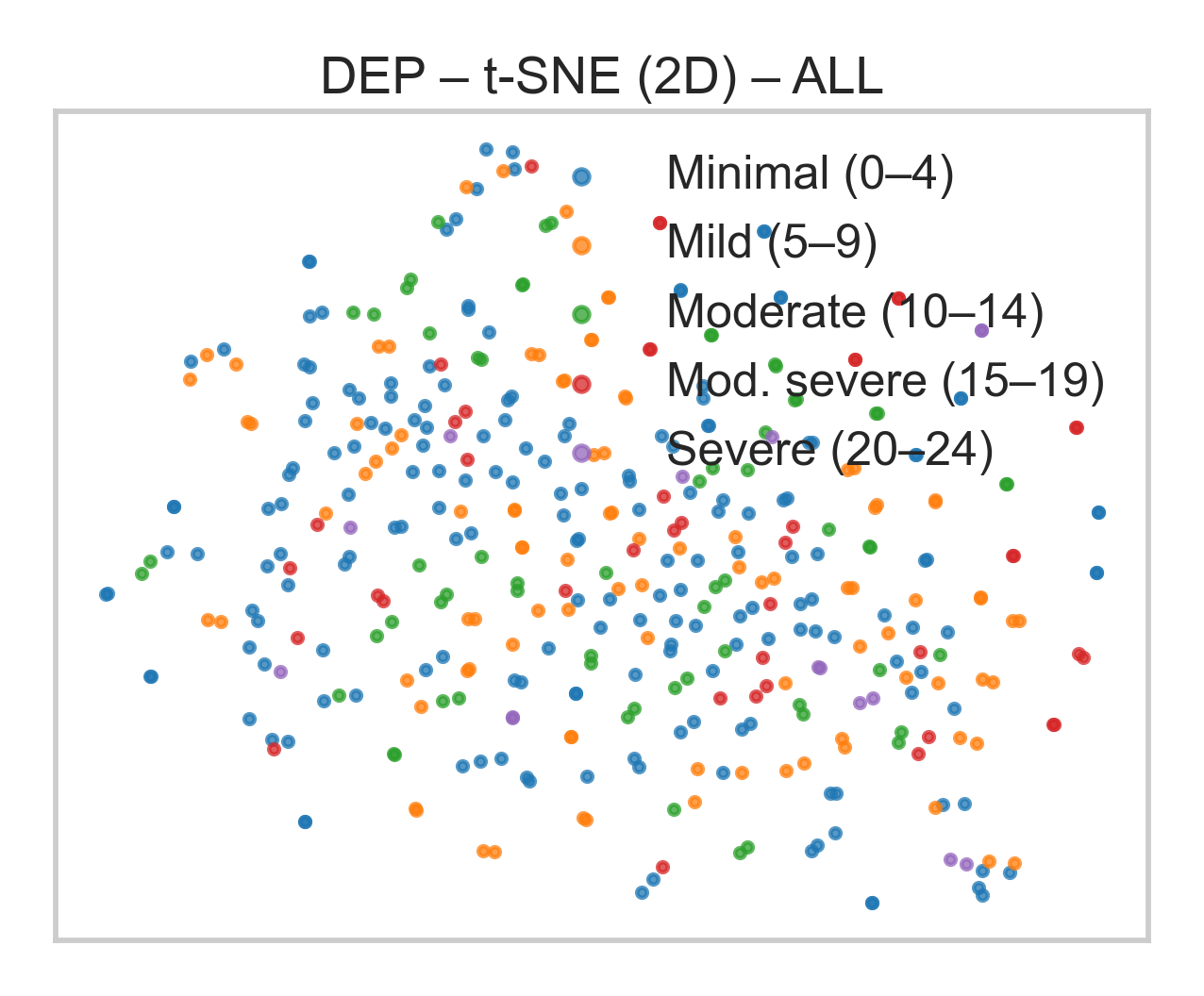}
    \subcaption{t-SNE — Depression}
    \label{fig:tsne-dep}
  \end{subfigure}\hfill
  \begin{subfigure}{0.48\linewidth}
    \centering
    \includegraphics[width=\linewidth]{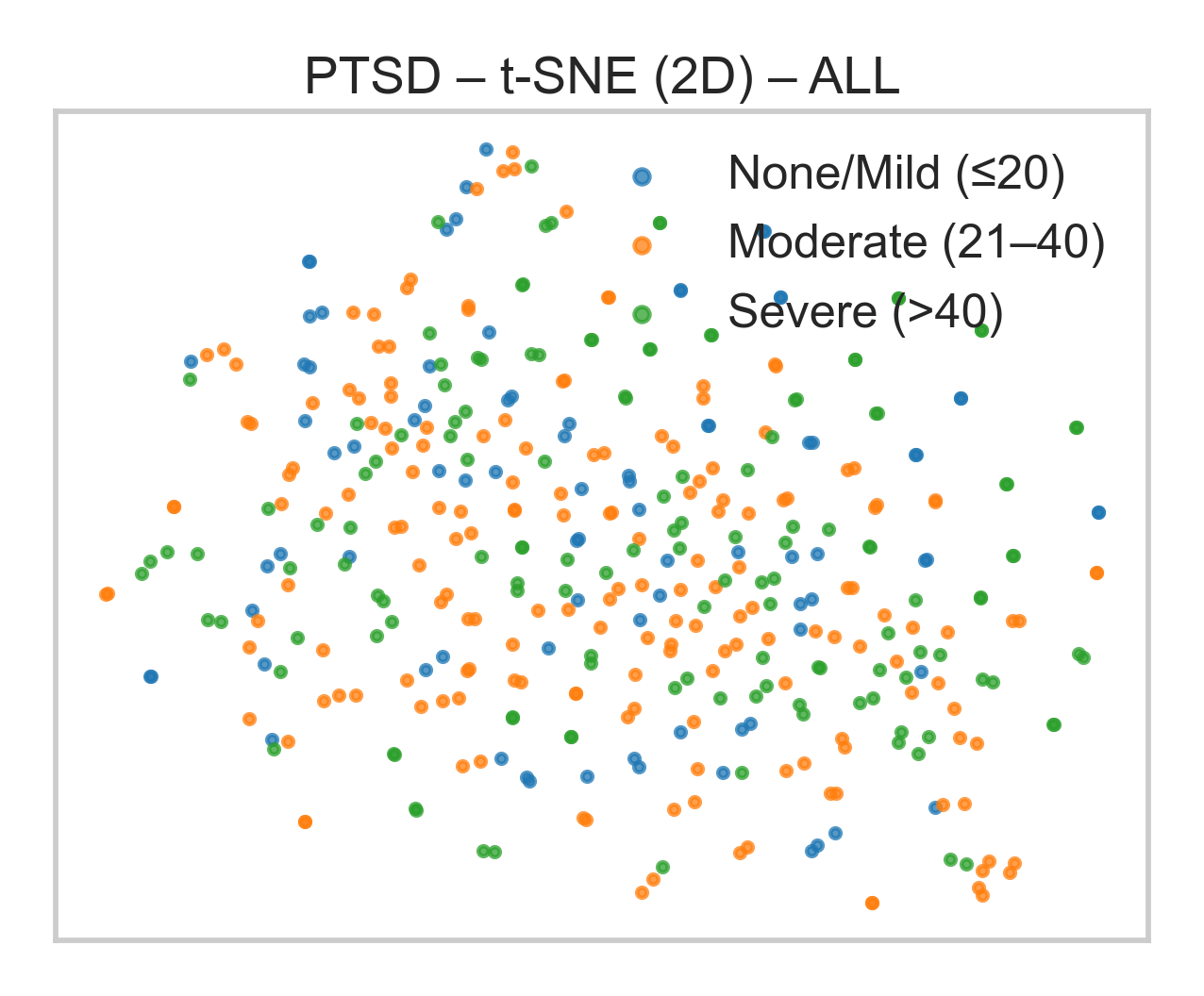}
    \subcaption{t-SNE — PTSD}
    \label{fig:tsne-ptsd}
  \end{subfigure}

  \caption{Embedding views of fused representations. Non-linear projections (t-SNE) reveal clearer cluster structure.}
  \label{fig:embeds}
\end{figure}

Figure~\ref{fig:embeds} illustrates fused embedding projections. PCA provides coarse class separation, but t-SNE reveals clearer cluster structures, especially for severe depression and PTSD. These visualizations corroborate the model’s ability to carve diagnostic structure in the latent space, though overlap persists in adjacent severity levels, consistent with the per-class F1 results.

\begin{figure}[t]
  \centering
  \begin{subfigure}[t]{.48\linewidth}
    \centering
    \includegraphics[width=\linewidth]{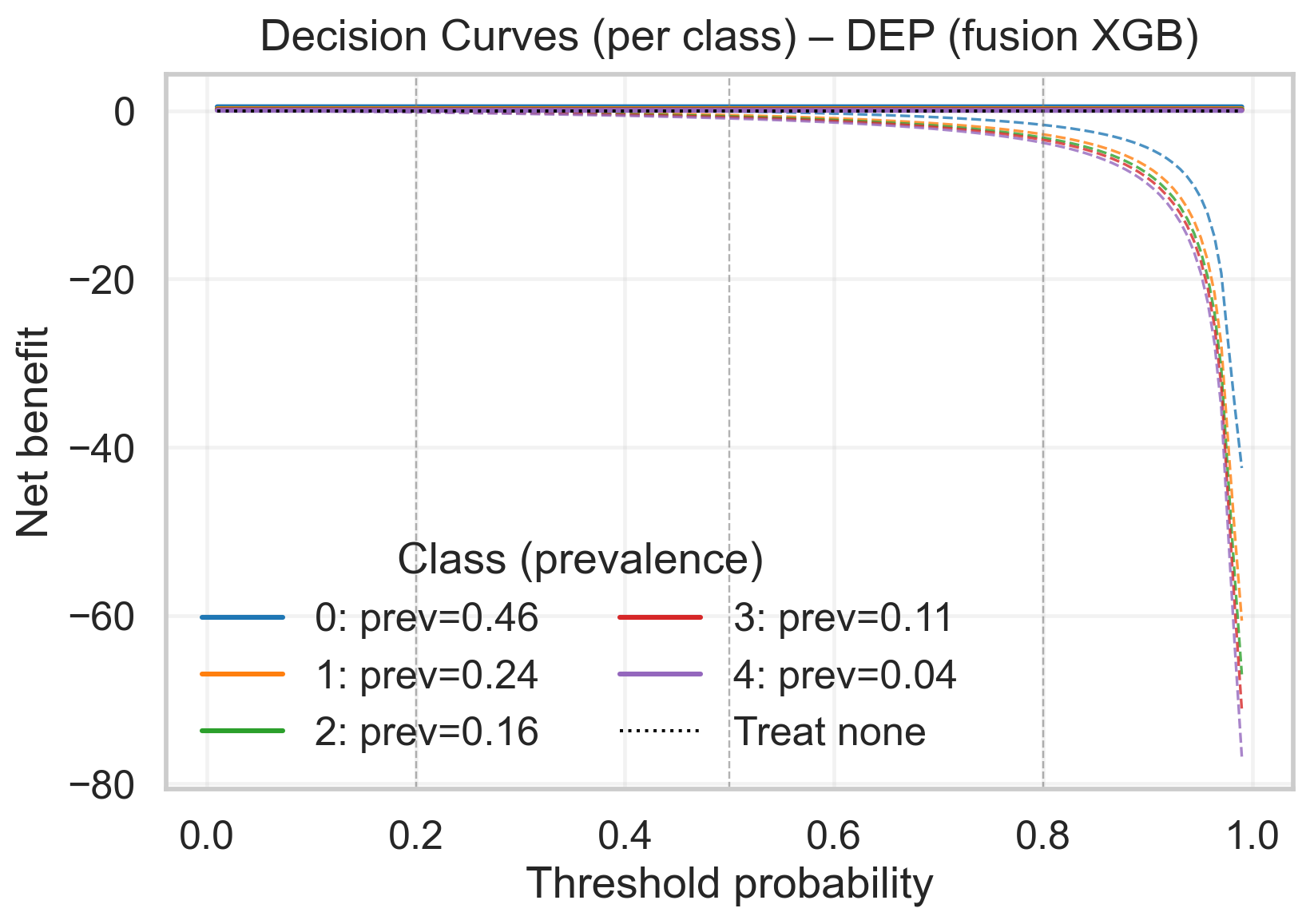}
    \caption{Depression}
  \end{subfigure}\hfill
  \begin{subfigure}[t]{.48\linewidth}
    \centering
    \includegraphics[width=\linewidth]{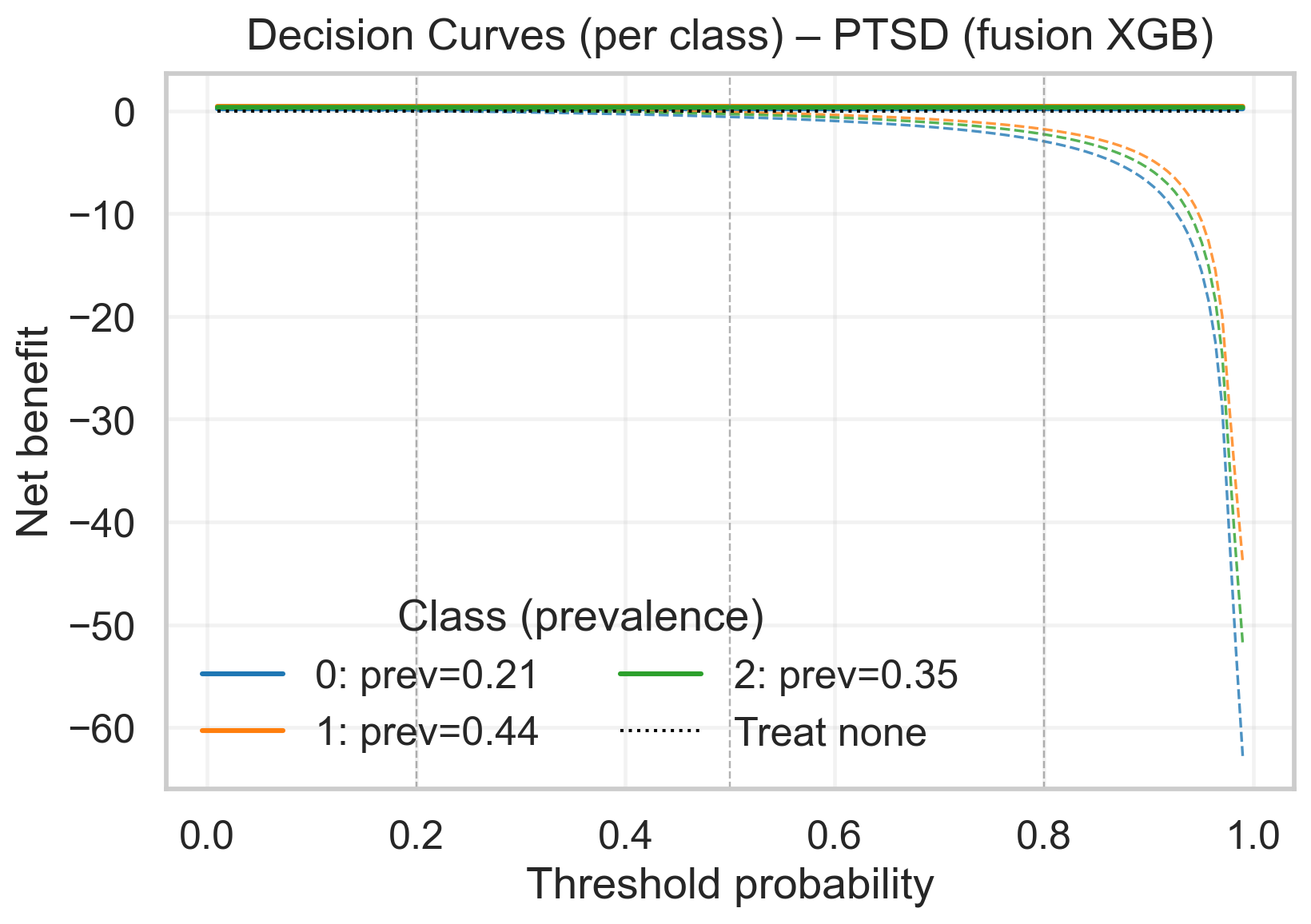}
    \caption{PTSD}
  \end{subfigure}
  \caption{Clinical decision analysis (net benefit vs.\ threshold). Fusion provides higher net benefit across a wide range of thresholds.}
  \label{fig:decision-curves}
\end{figure}

Beyond accuracy, clinical utility was examined through decision curves (Figure~\ref{fig:decision-curves}). Fusion models deliver higher net benefit across wide probability thresholds compared to unimodal baselines, confirming their relevance for decision support. For depression, net benefit remains consistently above ``treat all'' and ``treat none'' strategies, while PTSD curves show flatter slopes, reflecting greater class overlap and the challenge of intermediate severities.

Taken together, these results demonstrate that the proposed tri-modal system not only improves predictive performance but also aligns with clinical reasoning: extremes are easier to classify, middle severities remain difficult, and modality contributions vary by disorder. Fusion provides robustness against unimodal weaknesses, interpretability confirms clinically valid cues, and embedding visualizations validate the learned structure. This multi-angle evaluation underscores the framework’s readiness for translational use, while also highlighting the need for larger datasets and richer feature extraction to resolve mid-severity ambiguities.

\begin{figure}[t]
  \centering
  \begin{subfigure}[t]{.48\linewidth}
    \centering
    \includegraphics[width=\linewidth]{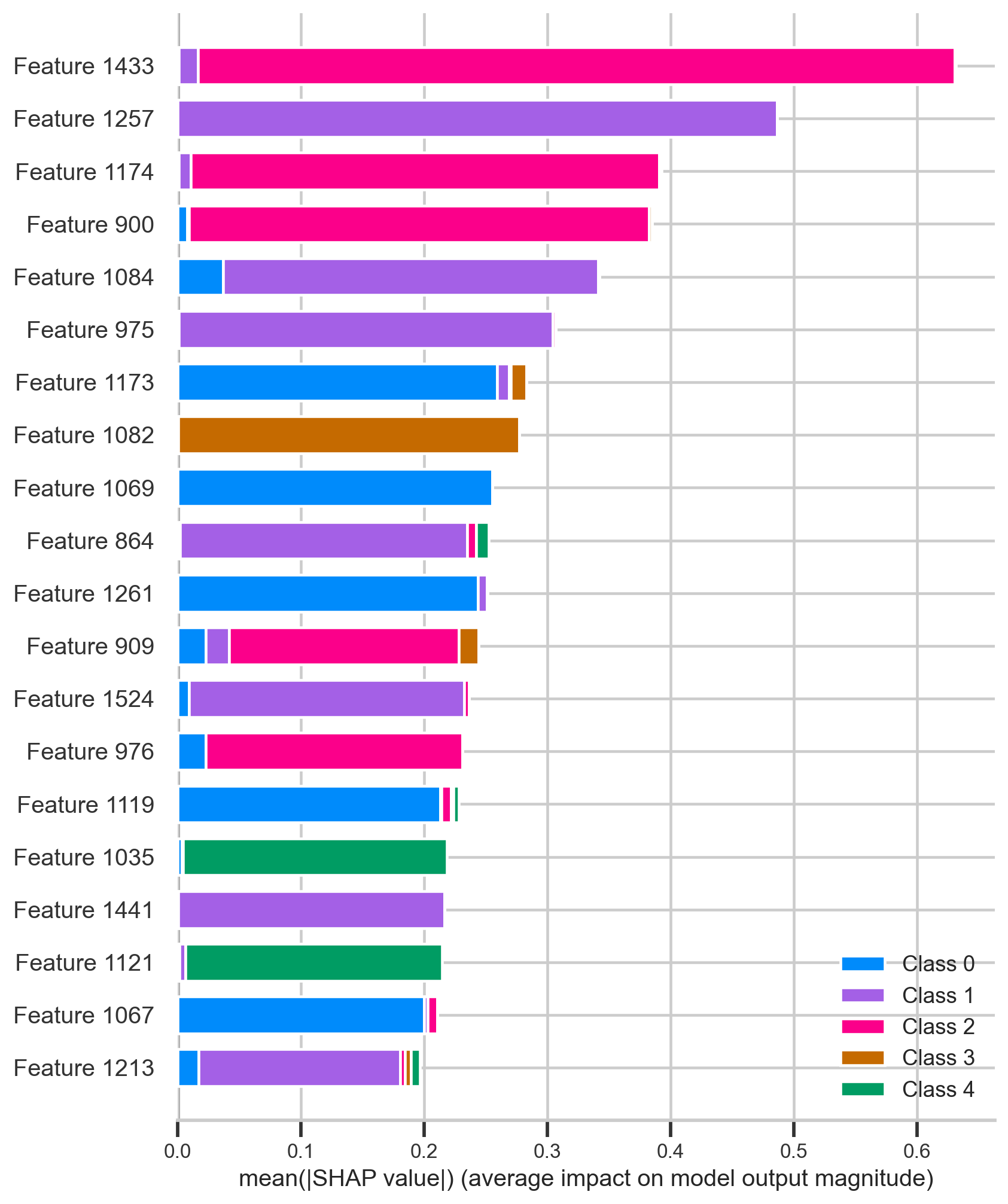}
    \caption{Depression SHAP summary (fusion)}
  \end{subfigure}\hfill
  \begin{subfigure}[t]{.48\linewidth}
    \centering
    \includegraphics[width=\linewidth]{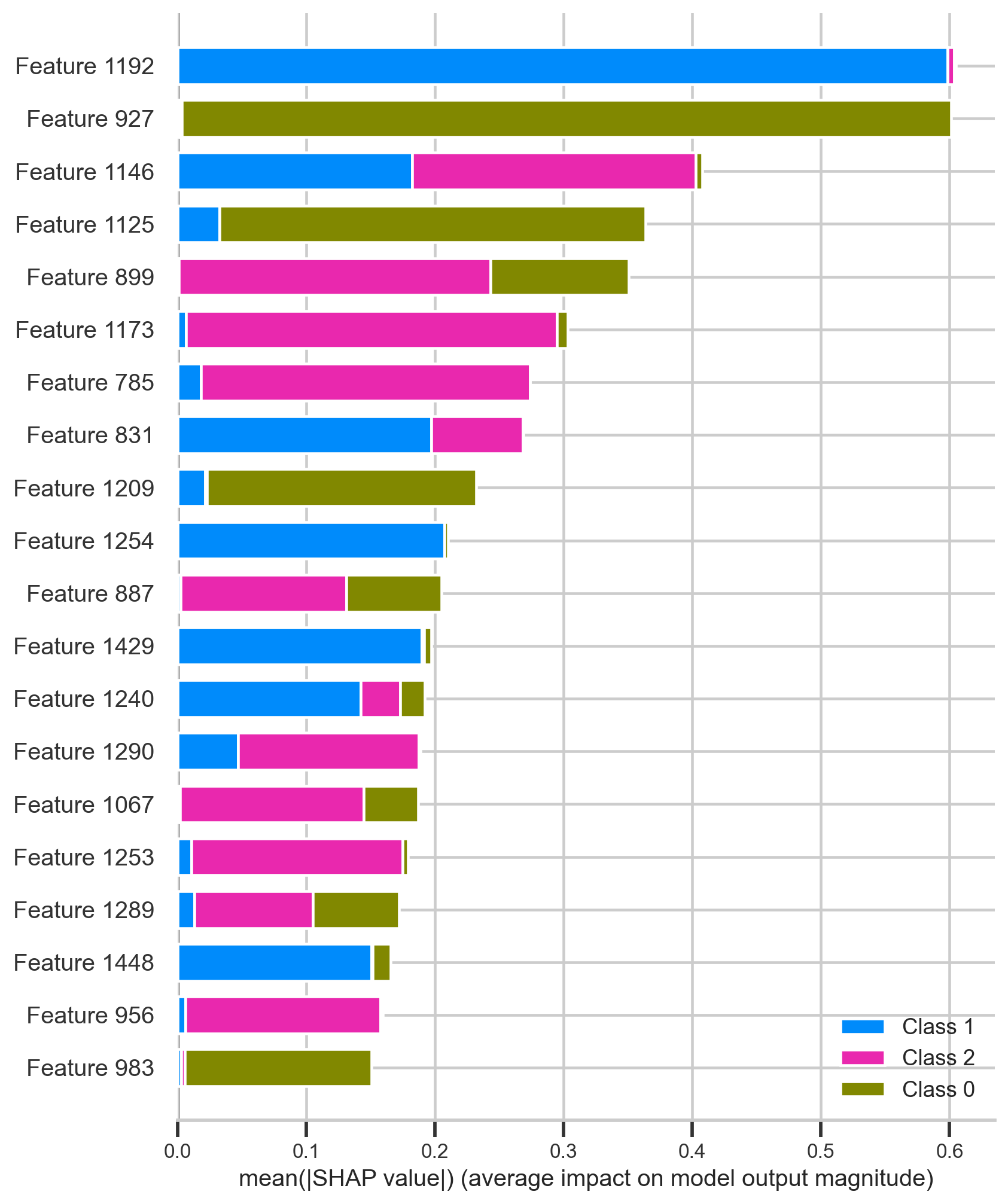}
    \caption{PTSD SHAP summary (fusion)}
  \end{subfigure}
  \caption{Model interpretability via SHAP. Top audio, facial, and text features contributing to decisions in the fusion XGB.}
  \label{fig:shap-fusion}
\end{figure}

\textbf{Clinical interpretability.} The resulting embeddings are not treated as opaque black-box signals. Through SHAP analysis (see Fig.~\ref{fig:shap-fusion}), we trace back highly weighted dimensions to linguistic phenomena such as increased first-person pronoun frequency, restricted vocabulary range, and polarity of sentiment-bearing tokens, all of which have been independently validated as markers of depression and PTSD. Thus, the text subsystem provides not only strong predictive accuracy but also interpretable signals aligned with established clinical theory, a prerequisite for trust calibration and clinical uptake. Linguistic markers (e.g., negative affect tokens, pronoun usage) dominate depression predictions, while prosodic cues (hesitation, jitter) and facial expressivity contribute more strongly to PTSD. This aligns with clinical knowledge: language reflects cognitive-emotional state in depression, whereas arousal-driven vocal and facial markers are salient for PTSD.

A consistent theme in the depression literature is that most systems are evaluated as \emph{binary} detectors (depressed vs.\ non-depressed), typically fusing audio and text and reporting very high discrimination when labels are collapsed to a single threshold. Al Hanai et al.\ modeled interview dynamics with dual LSTM branches COVAREP features for audio and Doc2Vec for text merging the streams in a feed-forward head to capture complementary, time-varying cues \cite{al2018detecting}. Lin et al.\ combined a BiLSTM+attention encoder for transcripts with a 1D-CNN for speech, feeding modality heads into parallel fully connected layers for classification; their architecture diagnoses presence and also \emph{assesses} severity, though results are reported mainly in terms of detection performance \cite{lin2020towards}. Lam et al.\ introduced a context-aware, data-driven pipeline that augments training data via topic modeling and then fuses deep 1D-CNN (audio) with Transformer-based text encoders, again showing that multimodal fusion outperforms unimodal branches \cite{lam2019context}.

Building on these trends, Jo and Kwak proposed a \emph{four-stream} BiLSTM/CNN model over audio and text feature families, emphasizing complementary acoustic descriptors (e.g., MFCC and GTCC) alongside richer linguistic processing \cite{jo2022diagnosis}. In their summary comparison (Table XIV), the bolded rows report the proposed model’s binary depression results: on \textbf{DAIC-WOZ}, \textbf{F1=0.97}, Precision=\textbf{0.97}, Recall=\textbf{0.97}; on \textbf{E-DAIC/``EDAIC''}, \textbf{F1=0.99}, Precision=\textbf{1.00}, Recall=\textbf{0.98}. These numbers substantially exceed earlier baselines they cite (e.g., F1=0.77–0.87 for prior DAIC-WOZ systems), reinforcing that careful multimodal fusion of audio+text can yield near-ceiling performance when the task is framed as a binary decision.

Our work targets a different—and arguably more clinically useful—problem formulation. Rather than binary screening, we predict *graded* severities (PHQ-8: 5 classes; PTSD: 3 classes) within a single tri-modal (text–audio–face) framework. Under stratified 5-fold cross-validation, the best models achieve Depression ACC$\approx$0.86 and F1\textsubscript{w}$\approx$0.86, and PTSD ACC$\approx$0.86 and F1\textsubscript{w}$\approx$0.86 (Table~\ref{tab:results_all}). These figures are not directly comparable to the 0.97–0.99 F1 reported by Jo and Kwak \cite{jo2022diagnosis}, whose task is binary; multi-class, severity-aware prediction is inherently harder because it must discriminate adjacent strata (e.g., mild vs.\ moderate vs.\ moderately severe). Their results represent the ceiling for binary detection on DAIC-style interviews, whereas ours show stable performance on a multi-class endpoint aligned with stepped-care decision points.

On PTSD, a closer, metric-matched comparison is possible. Dia et al.\ introduced a stochastic multimodal Transformer that explicitly models uncertainty for PTSD in EDAIC, reporting CCC = 0.722 with RMSE = 1.98 \cite{dia2024paying}. Our fusion system yields CCC = 0.743(higher is better) at RMSE = 2.03 (slightly worse), indicating improved agreement with the ground-truth distribution at essentially the same absolute error scale. Given that CCC is sensitive to both correlation and mean/variance alignment, the gain in concordance suggests our calibrated late-fusion probabilities better track clinical score variability, which is valuable for threshold-based triage and decision-curve analysis.

Finally, our design choices directly address several methodological gaps summarized by Jo and Kwak \cite{jo2022diagnosis}. They note that prior audio work often uses either 1D descriptors or 2D time–frequency maps in isolation and that text pipelines are frequently LSTM-centric, potentially limiting the capture of word-order and longer-range discourse structure. We explicitly: (i) combine prosodic log-Mel statistics (with deltas) that summarize time–frequency dynamics into a compact 256-D representation; (ii) use sentence-transformer embeddings for transcripts to encode semantics and pragmatics beyond local sequential cues; and (iii) add a \emph{third} visual stream (OpenFace-based 512-D descriptors) to capture nonverbal expressivity absent from the four-stream audio+text model. In ablations, removing text degrades performance the most, but audio and face consistently add net benefit, particularly for PTSD, supporting the premise that psychopathology manifests as coordinated verbal, vocal, and facial patterns.

\emph{In summary}, Jo and Kwak’s four-stream system \cite{jo2022diagnosis} sets a strong benchmark for \emph{binary} depression detection on DAIC/E-DAIC (F1 up to 0.99). Our contribution is orthogonal: we extend multimodal assessment to (a) \emph{severity-aware, multi-class} outputs for depression and PTSD within one pipeline, and (b) competitive PTSD agreement (
CCC
= 
0.743
CCC=0.743) relative to a recent stochastic Transformer \cite{dia2024paying}. Thus, while we do not surpass the current benchmark of 0.97–0.99 F1 values on \emph{their} binary endpoint, we advance clinical relevance by modeling multi-level severity across \emph{two} disorders and by demonstrating calibrated fusion that improves threshold-sensitive utility capabilities that prior audio+text systems \cite{al2018detecting,lin2020towards,lam2019context,jo2022diagnosis} did not target.

\section{Discussion}
\label{sec:Discussion}

\subsection{Research Implications}
\label{sec:discussion-research}
Our results demonstrate that framing mental-health assessment as a \emph{joint, multi-task} problem over disorders and \emph{graded severities} is both feasible and empirically beneficial. Rather than optimizing isolated, binary detectors, a shared representation with task-specific heads yields calibrated, decision-relevant probabilities for PHQ-8 and PTSD categories (cf.\ Section~\ref{sec:results}). This reframing invites new benchmarks where success is measured not only by discrimination, but by agreement with clinical severity strata and the consequent net benefit for thresholded decisions. Notably, when transcripts are unavailable or ASR is unreliable (phone triage), a bi-modal AUDIO+FACE model still attains viable accuracy (DEP/PTSD $\approx$0.83/0.82; Table~\ref{tab:results_all}), whereas text-only would be unusable tri-modal training yields the most robust, well-calibrated probabilities across such settings.

Ablations show that language dominates depression severity, while paralinguistic and nonverbal cues contribute disproportionately to PTSD differentiation. The systematic unimodal failure modes in the middle tiers (Fig.~\ref{fig:cm-unimodal-critical}) and the consistent gains of fusion (Fig.~\ref{fig:modality-bars}) imply that future architectures should \emph{explicitly} encode \emph{disorder–modality priors} (e.g., modality-aware gating, disorder-conditional attention) rather than treating streams as interchangeable. This also suggests curriculum strategies that first stabilize text signals for depression and then incrementally introduce audio/face synergies for PTSD.

Because severity thresholds underpin referral, stepped care, and safety escalations, well-behaved probabilities matter as much as raw accuracy. The observed decision-curve advantages of our fused, calibrated outputs indicate that \emph{calibration-aware training} (e.g., focal/ordinal losses with temperature or Dirichlet calibration) should become standard in computational psychiatry. Future work should report calibration diagnostics (ECE/MCE), class-conditional reliability diagrams, and net-benefit analyses alongside F1/AUC.

Although we report categorical metrics for comparability, the label structure is ordinal (PHQ-8 tiers) and correlated across disorders. This motivates (i) \emph{ordinal} objectives (cumulative link / CORN losses) to reduce mid-tier confusions; (ii) \emph{structured multi-task} heads that share uncertainty across disorders; and (iii) \emph{hierarchical} decoders (screening $\rightarrow$ coarse severity $\rightarrow$ fine severity) that match clinical triage flows.

Despite the field’s enthusiasm for very large, sequence-heavy models, our compact featurization + boosted late fusion delivered competitive PTSD concordance and robust per-class behavior while remaining easy to reproduce and audit. This argues for publishing \emph{lightweight, well-calibrated baselines} as first-class artefacts in mental-health ML, enabling fair comparison to heavier end-to-end systems.

Middle-severity ambiguity persists even with fusion, pointing to label noise and inter-rater variability. We see three priorities: (1) \emph{richer supervision} (ordinal + longitudinal labels, clinician rationales, anchor vignettes); (2) \emph{generalization protocols} (cross-site, cross-microphone, and demographically stratified evaluation with domain shift controls); and (3) \emph{uncertainty reporting} (predictive intervals at the \emph{patient} level) to support safe human-in-the-loop use. Together these will help convert strong offline metrics into trustworthy, real-world performance.

\subsection{Practical Implications}
\label{sec:discussion-practical}
Because outputs are severity-aligned rather than binary, the model can back clinical pathways that depend on thresholds (e.g., PHQ-8 $\geq$~\emph{moderate}) and jointly account for PTSD burden. Calibrated probabilities enable configurable operating points that match service capacity, with decision-curve analysis guiding threshold selection for screening versus case-finding contexts.

Late fusion with trust weighting tolerates missing or degraded modalities, a prerequisite for telehealth and community screening where microphones, cameras, or transcripts may be unreliable. In practice, this supports \emph{graceful degradation}: deploy text+audio in phone interviews, add face when video quality permits, and still return severity estimates with quantified confidence.

We recommend exposing (i) calibrated class probabilities per disorder; (ii) succinct, clinician-facing rationales (feature-level attributions aggregated to interpretable concepts); and (iii) confidence flags that trigger mandatory human review. These mechanics curb over-reliance, support shared decision-making, and fit existing triage and supervision models.

Given the sensitivity of AI for mental-health, deployments should include: pre-specified \emph{intended-use} statements; periodic \emph{calibration audits} on local populations; \emph{subgroup error analysis} with mitigation (recalibration or group-aware thresholds when justified); and \emph{incident response} protocols for misclassification in high-stakes cases (e.g., crisis pathways). Privacy-by-design pipelines—on-device feature extraction, ephemeral intermediates, and minimal data retention—are essential for acceptance.

The system’s compact footprint eases EHR or telehealth integration: transcripts from virtual consults feed the text encoder; audio/video is processed on edge or within secure clinical VPCs; and predictions flow back as structured fields with provenance. Post-deployment, institutions should maintain a \emph{living calibration layer} (temperature/Platt parameters) to track drift, plus routine shadow testing on clinician-labeled samples.

In resource-constrained services, multi-disorder, severity-aware triage can reduce unnecessary follow-ups for low-risk cases while prioritizing complex comorbidity. For longitudinal care, stable, calibrated trajectories support \emph{measurement-based practice}—flagging meaningful change rather than noise—without increasing clinician burden.

Near-term: ordinal losses, disorder-conditional gating, and richer rationales; mid-term: cross-site generalization studies with fairness audits; long-term: prospective trials embedding the tool in stepped-care pathways with outcome endpoints (remission, time-to-intervention), and governance artifacts (model card, calibration card, and data protection impact assessment) accompanying each deployment.

\section{Conclusion}
\label{sec:Conclusion}
  This paper introduced the first \emph{tri-modal, multi-disorder, severity-aware} diagnostic model for mental health, targeting both depression and PTSD within a unified framework. Unlike prior systems that either restrict themselves to binary screening or focus on a single disorder, our approach demonstrates that compact, reproducible fusion of text, audio, and face signals can deliver calibrated, severity-aligned predictions across comorbid conditions. By aligning outputs with validated clinical scales and evidencing decision-curve advantages, the model advances the field from mere detection toward actionable, stepped-care decision support.

Our results show that severity-aware, multi-task modeling is not only technically feasible but also yields reliable, trust-calibrated probabilities that map directly onto clinical thresholds. This represents a shift in evaluation standards: future benchmarks should not be satisfied with headline accuracy, but instead incorporate calibration, ordinal agreement, and net-benefit analysis as first-class metrics. In doing so, we move toward models that support clinical reasoning rather than merely classify.

The framework’s resilience to partial or degraded modalities and its transparent attribution of diagnostic cues make it viable for deployment in heterogeneous, real-world settings—whether telehealth consultations, outpatient triage, or longitudinal monitoring. Because predictions are severity-aware, they enable risk-stratified decision-making, prioritization of high-burden comorbid cases, and evidence-based follow-up planning. In combination with a transparent feature pipeline, strict participant-level CV, and a public code release, the system advances beyond a laboratory prototype toward an ethically aligned, practically usable tool for clinician-in-the-loop workflows.

The innovation here lies not only in modality fusion, but in explicitly tackling the two challenges that have limited clinical translation—multi-disorder overlap and severity granularity. Addressing these opens a new trajectory for research: developing disorder-conditional fusion mechanisms, ordinal and hierarchical objectives that respect symptom structure, and prospective trials embedding calibrated AI predictions into stepped-care pathways. By reframing automated mental health assessment in this way, we chart a course toward AI systems that are not just accurate but clinically actionable, trustworthy, and aligned with the complex realities of comorbid mental illness.

\section*{Declarations}

\textbf{Funding}    
This research was supported by the International Research Excellence Scholarship (iMQRES), Allocation Number: 20246179, funded by Macquarie University.

\textbf{Ethics approval and consent to participate}  
This study used only publicly available, de-identified corpora (DAIC-WOZ, E-DAIC). No new human data was collected; thus, no additional ethics approval was required.  

\textbf{Availability of data and materials}  
The datasets analyzed during the current study are publicly available:  
- DAIC-WOZ \& E-DAIC: \url{https://dcapswoz.ict.usc.edu}  

\textbf{Code availability}  
All code for feature extraction, training, and evaluation is available at: \url{https://github.com/cenacchi2000/EDAI-Trimodal/}  

\textbf{Authors' contributions}  
FC implemented the framework, ran experiments, and wrote the manuscript. DR supervised the project and revised the manuscript. LC contributed to theoretical framing and manuscript editing. All authors read and approved the final manuscript.

\bibliographystyle{IEEEtran}
\bibliography{sn-bibliography}

\begin{thebibliography}{10}
\providecommand{\url}[1]{#1}
\csname url@samestyle\endcsname
\providecommand{\newblock}{\relax}
\providecommand{\bibinfo}[2]{#2}
\providecommand{\BIBentrySTDinterwordspacing}{\spaceskip=0pt\relax}
\providecommand{\BIBentryALTinterwordstretchfactor}{4}
\providecommand{\BIBentryALTinterwordspacing}{\spaceskip=\fontdimen2\font plus
\BIBentryALTinterwordstretchfactor\fontdimen3\font minus \fontdimen4\font\relax}
\providecommand{\BIBforeignlanguage}[2]{{%
\expandafter\ifx\csname l@#1\endcsname\relax
\typeout{** WARNING: IEEEtran.bst: No hyphenation pattern has been}%
\typeout{** loaded for the language `#1'. Using the pattern for}%
\typeout{** the default language instead.}%
\else
\language=\csname l@#1\endcsname
\fi
#2}}
\providecommand{\BIBdecl}{\relax}
\BIBdecl

\bibitem{salleh2018burden}
M.~R. Salleh, ``The burden of mental illness: An emerging global disaster,'' \emph{Journal of Clinical and Health Sciences}, vol.~3, no.~1, pp. 1--8, 2018.

\bibitem{COVID192024global}
L.~Cao, \emph{GLOBAL COVID-19 RESEARCH AND MODELING: A Historical Record}.\hskip 1em plus 0.5em minus 0.4em\relax SPRINGER VERLAG, SINGAPOR, 2024.

\bibitem{sowers2019mental}
K.~M. Sowers, C.~N. Dulmus, and B.~K. Linn, ``Mental illness: worldwide,'' in \emph{Encyclopedia of Social Work}, 2019.

\bibitem{yan2024global}
G.~Yan, Y.~Zhang, S.~Wang, Y.~Yan, M.~Liu, M.~Tian, and W.~Tian, ``Global, regional, and national temporal trend in burden of major depressive disorder from 1990 to 2019: An analysis of the global burden of disease study,'' \emph{Psychiatry research}, vol. 337, p. 115958, 2024.

\bibitem{hoang2023prevalence}
V.~T.~H. Hoang, T.~N.~T. Huyen, N.~D. Thi, G.~L. Minh \emph{et~al.}, ``Prevalence of post-traumatic stress disorder in general population during covid-19 pandemic: An umbrella review and meta-analysis,'' \emph{OBM Neurobiology}, vol.~7, no.~3, pp. 1--15, 2023.

\bibitem{abdullah2022prevalence}
R.~Abdullah, A.~Suryoputro, and M.~Sakundarno, ``The prevalence of post-traumatic stress disorder on healthcare workers during the covid-19 pandemic: Literature review,'' \emph{GUIDENA: Jurnal Ilmu Pendidikan, Psikologi, Bimbingan dan Konseling}, vol.~12, no.~3, pp. 341--349, 2022.

\bibitem{lei2021prevalence}
L.~Lei, H.~Zhu, Y.~Li, T.~Dai, S.~Zhao, X.~Zhang, X.~Muchu, and S.~Su, ``Prevalence of post-traumatic stress disorders and associated factors one month after the outbreak of the covid-19 among the public in southwestern china: a cross-sectional study,'' \emph{BMC Psychiatry}, vol.~21, no.~1, p. 545, 2021.

\bibitem{field2024post}
T.~Field, ``Post-traumatic stress disorder research: a narrative review,'' \emph{Journal of Psychology \& Clinical Psychiatry}, vol.~15, no.~6, pp. 303--307, 2024.

\bibitem{miller2024quality}
C.~R. Miller, J.~E. McDonald, P.~P. Grau, and C.~T. Wetterneck, ``Quality of life in posttraumatic stress disorder: The role of posttraumatic anhedonia and depressive symptoms in a treatment-seeking community sample,'' \emph{Trauma Care}, vol.~4, no.~1, pp. 1--14, 2024.

\bibitem{kline2024ptsd}
A.~C. Kline, N.~Otis, K.~E. Panza, C.~T. McCabe, L.~Glassman, J.~S. Campbell, and K.~H. Walter, ``Ptsd, depression, and treatment outcomes: A latent profile analysis among active duty personnel in a residential ptsd program,'' \emph{Journal of psychiatric research}, vol. 173, pp. 71--79, 2024.

\bibitem{brigido2021posttraumatic}
S.~Brigido, M.~Bozzay, and N.~S. Philip, ``Posttraumatic stress disorder symptom severity does not predict depression improvement, but may impact clinical response and remission,'' \emph{The Journal of clinical psychiatry}, vol.~82, no.~3, p. 30701, 2021.

\bibitem{pejuskovic2020longitudinal}
B.~Pejuskovic, D.~Lecic-Tosevski, and O.~Toskovic, ``Longitudinal study of ptsd and depression in a war-exposed sample--comorbidity increases distress and suicide risk,'' \emph{Global Psychiatry Archives}, vol.~3, no.~1, pp. 64--71, 2020.

\bibitem{radell2020depression}
M.~L. Radell, E.~A. Hamza, and A.~A. Moustafa, ``Depression in post-traumatic stress disorder,'' \emph{Reviews in the Neurosciences}, vol.~31, no.~7, pp. 703--722, 2020.

\bibitem{pejuvskovic2024posttraumatic}
B.~Peju{\v{s}}kovi{\'c}, ``Posttraumatic stress disorder--an overview in new diagnosis and treatment approaches,'' \emph{MEDICINSKA IS RA I AN A}, p.~75, 2024.

\bibitem{schultebraucks2022deep}
K.~Schultebraucks, V.~Yadav, A.~Y. Shalev, G.~A. Bonanno, and I.~R. Galatzer-Levy, ``Deep learning-based classification of posttraumatic stress disorder and depression following trauma utilizing visual and auditory markers of arousal and mood,'' \emph{Psychological Medicine}, vol.~52, no.~5, pp. 957--967, 2022.

\bibitem{uddin2022deep}
M.~A. Uddin, J.~B. Joolee, and K.-A. Sohn, ``Deep multi-modal network based automated depression severity estimation,'' \emph{IEEE transactions on affective computing}, vol.~14, no.~3, pp. 2153--2167, 2022.

\bibitem{konig2022detecting}
A.~K{\"o}nig, J.~Tr{\"o}ger, E.~Mallick, M.~Mina, N.~Linz, C.~Wagnon, J.~Karbach, C.~Kuhn, and J.~Peter, ``Detecting subtle signs of depression with automated speech analysis in a non-clinical sample,'' \emph{BMC psychiatry}, vol.~22, no.~1, p. 830, 2022.

\bibitem{quatieri2023emotion}
T.~F. Quatieri, J.~Wang, J.~R. Williamson, R.~DeLaura, T.~Talkar, N.~P. Solomon, S.~E. Kuchinsky, M.~Eitel, T.~Brickell, S.~Lippa \emph{et~al.}, ``An emotion-driven vocal biomarker-based ptsd screening tool,'' \emph{IEEE Open Journal of Engineering in Medicine and Biology}, vol.~5, pp. 621--626, 2023.

\bibitem{dhelim2023artificial}
S.~Dhelim, L.~Chen, H.~Ning, and C.~Nugent, ``Artificial intelligence for suicide assessment using audiovisual cues: a review,'' \emph{Artificial Intelligence Review}, vol.~56, no.~6, pp. 5591--5618, 2023.

\bibitem{al2018detecting}
T.~Al~Hanai, M.~M. Ghassemi, and J.~R. Glass, ``Detecting depression with audio/text sequence modeling of interviews.'' in \emph{Interspeech}, 2018, pp. 1716--1720.

\bibitem{lin2020towards}
L.~Lin, X.~Chen, Y.~Shen, and L.~Zhang, ``Towards automatic depression detection: A bilstm/1d cnn-based model,'' \emph{Applied Sciences}, vol.~10, no.~23, p. 8701, 2020.

\bibitem{lam2019context}
G.~Lam, H.~Dongyan, and W.~Lin, ``Context-aware deep learning for multi-modal depression detection,'' in \emph{ICASSP 2019-2019 IEEE international conference on acoustics, speech and signal processing (ICASSP)}.\hskip 1em plus 0.5em minus 0.4em\relax IEEE, 2019, pp. 3946--3950.

\bibitem{jo2022diagnosis}
A.-H. Jo and K.-C. Kwak, ``Diagnosis of depression based on four-stream model of bi-lstm and cnn from audio and text information,'' \emph{IEEE Access}, vol.~10, pp. 134\,113--134\,135, 2022.

\bibitem{zhang2024multimodal}
Z.~Zhang, S.~Zhang, D.~Ni, Z.~Wei, K.~Yang, S.~Jin, G.~Huang, Z.~Liang, L.~Zhang, L.~Li \emph{et~al.}, ``Multimodal sensing for depression risk detection: Integrating audio, video, and text data,'' \emph{Sensors}, vol.~24, no.~12, p. 3714, 2024.

\bibitem{chen2024iifdd}
J.~Chen, Y.~Hu, Q.~Lai, W.~Wang, J.~Chen, H.~Liu, G.~Srivastava, A.~K. Bashir, and X.~Hu, ``Iifdd: Intra and inter-modal fusion for depression detection with multi-modal information from internet of medical things,'' \emph{Information Fusion}, vol. 102, p. 102017, 2024.

\bibitem{li2024fpt}
Y.~Li, X.~Yang, M.~Zhao, Z.~Wang, Y.~Yao, W.~Qian, and S.~Qi, ``Fpt-former: A flexible parallel transformer of recognizing depression by using audiovisual expert-knowledge-based multimodal measures,'' \emph{International Journal of Intelligent Systems}, vol. 2024, no.~1, p. 1564574, 2024.

\bibitem{dia2024paying}
M.~Dia, G.~Khodabandelou, and A.~Othmani, ``Paying attention to uncertainty: A stochastic multimodal transformers for post-traumatic stress disorder detection using video,'' \emph{Computer Methods and Programs in Biomedicine}, vol. 257, p. 108439, 2024.

\bibitem{gratch2014distress}
J.~Gratch, R.~Artstein, G.~M. Lucas, G.~Stratou, S.~Scherer, A.~Nazarian, R.~Wood, J.~Boberg, D.~DeVault, S.~Marsella \emph{et~al.}, ``The distress analysis interview corpus of human and computer interviews.'' in \emph{LREC}, vol.~14.\hskip 1em plus 0.5em minus 0.4em\relax Reykjavik, 2014, pp. 3123--3128.

\bibitem{ringeval2019avec}
F.~Ringeval, B.~Schuller, M.~Valstar, N.~Cummins, R.~Cowie, L.~Tavabi, M.~Schmitt, S.~Alisamir, S.~Amiriparian, E.-M. Messner \emph{et~al.}, ``Avec 2019 workshop and challenge: state-of-mind, detecting depression with ai, and cross-cultural affect recognition,'' in \emph{Proceedings of the 9th International on Audio/visual Emotion Challenge and Workshop}, 2019, pp. 3--12.

\bibitem{graham2019artificial}
S.~Graham, C.~Depp, E.~E. Lee, C.~Nebeker, X.~Tu, H.-C. Kim, and D.~V. Jeste, ``Artificial intelligence for mental health and mental illnesses: an overview,'' \emph{Current psychiatry reports}, vol.~21, no.~11, p. 116, 2019.

\bibitem{Gerantia_2024}
M.~Gerantia, ``Artificial intelligence in psychiatry: A comprehensive literature review,'' \emph{European Psychiatry}, vol.~67, no.~S1, p. S61–S61, 2024.

\bibitem{zhang2024can}
Z.~Zhang and J.~Wang, ``Can ai replace psychotherapists? exploring the future of mental health care,'' \emph{Frontiers in psychiatry}, vol.~15, p. 1444382, 2024.

\bibitem{zimmerman2024value}
M.~Zimmerman, ``The value and limitations of self-administered questionnaires in clinical practice and epidemiological studies,'' \emph{World Psychiatry}, vol.~23, no.~2, p. 210, 2024.

\bibitem{linden2012standardized}
M.~Linden and B.~Muschalla, ``Standardized diagnostic interviews, criteria, and algorithms for mental disorders: garbage in, garbage out,'' \emph{European archives of psychiatry and clinical neuroscience}, vol. 262, no.~6, pp. 535--544, 2012.

\bibitem{finlay2001methodological}
W.~M. Finlay and E.~Lyons, ``Methodological issues in interviewing and using self-report questionnaires with people with mental retardation.'' \emph{Psychological assessment}, vol.~13, no.~3, p. 319, 2001.

\bibitem{magruder2015diagnostic}
K.~Magruder, D.~Yeager, J.~Goldberg, C.~Forsberg, B.~Litz, V.~Vaccarino, M.~Friedman, T.~Gleason, G.~Huang, and N.~Smith, ``Diagnostic performance of the ptsd checklist and the vietnam era twin registry ptsd scale,'' \emph{Epidemiology and psychiatric sciences}, vol.~24, no.~5, pp. 415--422, 2015.

\bibitem{north2021symptom}
C.~S. North and D.~Baron, ``The symptom structure of postdisaster major depression: Convergence of evidence from 11 disaster studies using consistent methods,'' \emph{Behavioral Sciences}, vol.~11, no.~1, p.~8, 2021.

\bibitem{menne2024voice}
F.~Menne, F.~D{\"o}rr, J.~Schr{\"a}der, J.~Tr{\"o}ger, U.~Habel, A.~K{\"o}nig, and L.~Wagels, ``The voice of depression: speech features as biomarkers for major depressive disorder,'' \emph{BMC psychiatry}, vol.~24, no.~1, p. 794, 2024.

\bibitem{pandey2022deep}
S.~K. Pandey, H.~S. Shekhawat, S.~Prasanna, S.~Bhasin, and R.~Jasuja, ``A deep tensor-based approach for automatic depression recognition from speech utterances,'' \emph{Plos one}, vol.~17, no.~8, p. e0272659, 2022.

\bibitem{zhang2022_fusion_eeg}
\BIBentryALTinterwordspacing
B.~Zhang, D.~Wei, W.~Chang, Z.~Yang, and Y.~Li, ``Feature-level fusion based on spatial-temporal of pervasive eeg for depression recognition,'' \emph{Computer Methods and Programs in Biomedicine}, vol. 226, p. 107113, 2022. [Online]. Available: \url{https://pubmed.ncbi.nlm.nih.gov/36103735/}
\BIBentrySTDinterwordspacing

\bibitem{yousufi2024_eeg_audio}
\BIBentryALTinterwordspacing
M.~Yousufi \emph{et~al.}, ``Multimodal fusion of eeg and audio spectrogram for major depressive disorder classification,'' \emph{Brain Sciences}, vol.~14, no.~10, p. 1018, 2024. [Online]. Available: \url{https://www.mdpi.com/2076-3425/14/10/1018}
\BIBentrySTDinterwordspacing

\bibitem{wei2022_subattn}
\BIBentryALTinterwordspacing
X.~Wei \emph{et~al.}, ``Multi-modal depression estimation based on sub-attentional fusion,'' \emph{arXiv preprint}, 2022. [Online]. Available: \url{https://arxiv.org/abs/2207.06180}
\BIBentrySTDinterwordspacing

\bibitem{nykoniuk2024}
\BIBentryALTinterwordspacing
B.~Nykoniuk \emph{et~al.}, ``Multimodal data fusion for depression detection approach,'' \emph{Computers}, vol.~13, no.~1, p.~9, 2024. [Online]. Available: \url{https://www.mdpi.com/2079-3197/13/1/9}
\BIBentrySTDinterwordspacing

\bibitem{sadeghi2024}
\BIBentryALTinterwordspacing
E.~Sadeghi \emph{et~al.}, ``Harnessing multimodal approaches for depression severity prediction using the e-daic dataset,'' \emph{JMIR Mental Health}, vol.~11, no.~5, p. e53068, 2024. [Online]. Available: \url{https://pmc.ncbi.nlm.nih.gov/articles/PMC11666580/}
\BIBentrySTDinterwordspacing

\bibitem{xu2025}
\BIBentryALTinterwordspacing
Z.~Xu \emph{et~al.}, ``Depression detection methods based on multimodal fusion of voice and text,'' \emph{Scientific Reports}, vol.~15, no.~1, p. 3524, 2025. [Online]. Available: \url{https://www.nature.com/articles/s41598-025-03524-4}
\BIBentrySTDinterwordspacing

\bibitem{hiquE2024}
\BIBentryALTinterwordspacing
T.~Jiang \emph{et~al.}, ``Hique: Hierarchical question embedding network for multimodal depression detection,'' \emph{arXiv preprint}, 2024. [Online]. Available: \url{https://arxiv.org/abs/2408.03648}
\BIBentrySTDinterwordspacing

\bibitem{bucur2023s}
A.-M. Bucur, A.~Cosma, P.~Rosso, and L.~P. Dinu, ``It’s just a matter of time: Detecting depression with time-enriched multimodal transformers,'' in \emph{European conference on information retrieval}.\hskip 1em plus 0.5em minus 0.4em\relax Springer, 2023, pp. 200--215.

\bibitem{prama2024ai}
T.~T. Prama, M.~S. Islam, M.~M. Anwar, and I.~Jahan, ``Ai-enabled deep depression detection and evaluation informed by dsm-5-tr,'' \emph{IEEE Transactions on Computational Social Systems}, vol.~11, no.~5, pp. 6453--6465, 2024.

\bibitem{mendes2022sensing}
J.~P. Mendes, I.~R. Moura, P.~Van~de Ven, D.~Viana, F.~J. Silva, L.~R. Coutinho, S.~Teixeira, J.~J. Rodrigues, and A.~S. Teles, ``Sensing apps and public data sets for digital phenotyping of mental health: systematic review,'' \emph{Journal of medical Internet research}, vol.~24, no.~2, p. e28735, 2022.

\bibitem{wang2024application}
J.~Wang, H.~Ouyang, R.~Jiao, S.~Cheng, H.~Zhang, Z.~Shang, Y.~Jia, W.~Yan, L.~Wu, and W.~Liu, ``The application of machine learning techniques in posttraumatic stress disorder: a systematic review and meta-analysis,'' \emph{NPJ Digital Medicine}, vol.~7, no.~1, p. 121, 2024.

\bibitem{gagnon2022identifying}
P.~Gagnon-Sanschagrin, J.~Schein, A.~Urganus, E.~Serra, Y.~Liang, P.~Musingarimi, M.~Cloutier, A.~Gu{\'e}rin, and L.~L. Davis, ``Identifying individuals with undiagnosed post-traumatic stress disorder in a large united states civilian population--a machine learning approach,'' \emph{BMC psychiatry}, vol.~22, no.~1, p. 630, 2022.

\bibitem{quillivic2024interdisciplinary}
R.~Quillivic, F.~Gayraud, Y.~Aux{\'e}m{\'e}ry, L.~Vanni, D.~Peschanski, F.~Eustache, J.~Dayan, and S.~Mesmoudi, ``Interdisciplinary approach to identify language markers for post-traumatic stress disorder using machine learning and deep learning,'' \emph{Scientific reports}, vol.~14, no.~1, p. 12468, 2024.

\bibitem{song2020mpnet}
K.~Song, X.~Tan, T.~Qin, J.~Lu, and T.-Y. Liu, ``Mpnet: Masked and permuted pre-training for language understanding,'' in \emph{Advances in Neural Information Processing Systems 33 (NeurIPS)}, 2020, pp. 16\,857--16\,867.

\bibitem{reimers2019sentencebert}
N.~Reimers and I.~Gurevych, ``Sentence-bert: Sentence embeddings using siamese bert-networks,'' in \emph{Proceedings of the 2019 Conference on Empirical Methods in Natural Language Processing and the 9th International Joint Conference on Natural Language Processing (EMNLP-IJCNLP)}.\hskip 1em plus 0.5em minus 0.4em\relax Association for Computational Linguistics, 2019, pp. 3982--3992.

\bibitem{mollahosseini2019}
A.~Mollahosseini, B.~Hasani, and M.~H. Mahoor, ``Affectnet: A database for facial expression, valence, and arousal computing in the wild,'' \emph{IEEE Transactions on Affective Computing}, vol.~10, no.~1, pp. 18--31, 2019.

\bibitem{wang2024cmpb}
X.~Li, X.~Yi, J.~Ye, Y.~Zheng, and Q.~Wang, ``Sftnet: A microexpression-based method for depression detection,'' \emph{Computer Methods and Programs in Biomedicine}, vol. 243, p. 107923, 2024.

\bibitem{chen2016xgboost}
T.~Chen and C.~Guestrin, ``Xgboost: A scalable tree boosting system,'' in \emph{Proceedings of the 22nd ACM SIGKDD International Conference on Knowledge Discovery and Data Mining (KDD)}.\hskip 1em plus 0.5em minus 0.4em\relax ACM, 2016, pp. 785--794.

\bibitem{lundberg2017unified}
S.~M. Lundberg and S.-I. Lee, ``A unified approach to interpreting model predictions,'' in \emph{Advances in Neural Information Processing Systems 30 (NeurIPS)}, 2017, pp. 4765--4774.

\bibitem{kroenke2009phq8}
\BIBentryALTinterwordspacing
K.~Kroenke, T.~W. Strine, R.~L. Spitzer, J.~B.~W. Williams, J.~T. Berry, and A.~H. Mokdad, ``The phq-8 as a measure of current depression in the general population,'' \emph{Journal of Affective Disorders}, vol. 114, no. 1-3, pp. 163--173, 2009. [Online]. Available: \url{https://doi.org/10.1016/j.jad.2008.06.026}
\BIBentrySTDinterwordspacing

\bibitem{blevins2015pcl5}
\BIBentryALTinterwordspacing
C.~A. Blevins, F.~W. Weathers, M.~T. Davis, T.~K. Witte, and J.~L. Domino, ``The posttraumatic stress disorder checklist for dsm-5 (pcl-5): Development and initial psychometric evaluation,'' \emph{Journal of Traumatic Stress}, vol.~28, no.~6, pp. 489--498, 2015. [Online]. Available: \url{https://doi.org/10.1002/jts.22059}
\BIBentrySTDinterwordspacing

\end{thebibliography}

\providecommand{\bioimage}[1]{%
  \includegraphics[width=1.0in,height=1.45in,clip,keepaspectratio]{#1}%
}

\begin{IEEEbiography}[{\vspace{-2mm}\bioimage{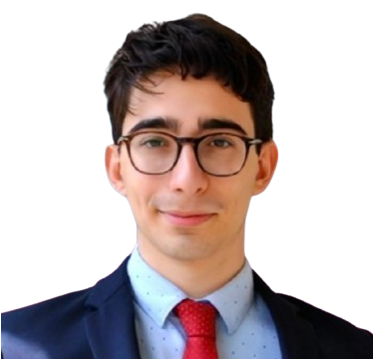}}]{Filippo Cenacchi} received the bachelor’s degree in Computer Science from the Free University of Bozen-Bolzano, Italy, and the M.Sc. degrees in Human–Computer Interaction from KTH Royal Institute of Technology, Sweden, and in Situated Interaction from Université Paris-Saclay, France. He is currently working toward the PhD degree in Computer Science with Macquarie University, Sydney. His research interests include HCI, VR/AR and humanoid AI.
\end{IEEEbiography}

\begin{IEEEbiography}[{\vspace{-2mm}\bioimage{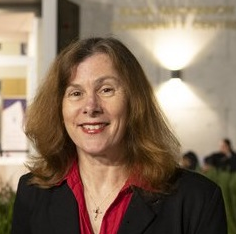}}]{Deborah Richards}
is a Professor in the School of Computing at Macquarie University, Sydney. She completed a PhD in artificial intelligence at the University of New South Wales and joined academia in 1999. Her current focus is on intelligent virtual agents, virtual worlds and serious games. With 20 years in industry prior to joining academia, her research is applied and interdisciplinary, and focused on appropriate and ethical use of technology particularly for health, wellbeing, education and training.
\end{IEEEbiography}


\begin{IEEEbiography}[{\includegraphics[width=1in,height=1.25in,clip,keepaspectratio]{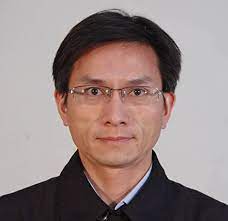}}]{Longbing Cao} (@SM in 2006) received a PhD degree in Pattern Recognition and Intelligent Systems from the Chinese Academy of Sciences, China, and another PhD degree in Computing Sciences at the University of Technology Sydney, Australia. He is the Distinguished Chair in Artificial Intelligence, Director of Frontier AI Research Centre, and an ARC Future Fellow (professorial level) at Macquarie University. His research covers AI, data science, machine learning,  behavior informatics, and their enterprise innovation.
\end{IEEEbiography}

\vfill

\end{document}